\documentclass[10pt,twocolumn,letterpaper]{article}

\usepackage{cvpr}              % To produce the CAMERA-READY version
\definecolor{cvprblue}{rgb}{0.21,0.49,0.74}
\usepackage[pagebackref,breaklinks,colorlinks,allcolors=cvprblue]{hyperref}

\usepackage{multirow}
\usepackage{makecell}
\usepackage{graphicx}
\def\paperID{248} % *** Enter the Paper ID here
\def\confName{CVPR}
\def\confYear{2026}

\title{%TokenSplat: Token-aligned Feed-forward 3D Gaussian Splatting from Unposed Image Sequences
TokenSplat: Token-aligned 3D Gaussian Splatting for Feed-forward\\Pose-free Reconstruction
}

\author{
    Yihui Li\textsuperscript{\rm 1,2},
    Chengxin Lv\textsuperscript{\rm 1,2},
    Zichen Tang\textsuperscript{\rm 1,3},
    Hongyu Yang\textsuperscript{\rm 3*},
    Di Huang\textsuperscript{\rm 1,2} \\
    \textsuperscript{\rm 1} State Key Laboratory of Complex and Critical Software Environment, Beijing, China \\
    \textsuperscript{\rm 2} School of Computer Science and Engineering, Beihang University, China\\
    \textsuperscript{\rm 3} School of Artificial Intelligence, Beihang University, China\\
    {\tt\small \{kidleyh, chengxinlv, zctang, hongyuyang, dhuang\}@buaa.edu.cn}
}
\begin{document}
\maketitle
\begin{abstract}
We present \textbf{TokenSplat}, a feed-forward framework for joint 3D Gaussian reconstruction and camera pose estimation from unposed multi-view images.
At its core, TokenSplat introduces a \textbf{Token-aligned Gaussian Prediction} module that aligns semantically corresponding information across views directly in the feature space.
Guided by coarse token positions and fusion confidence, it aggregates multi-scale contextual features to enable long-range cross-view reasoning and reduce redundancy from overlapping Gaussians.
To further enhance pose robustness and disentangle viewpoint cues from scene semantics, TokenSplat employs learnable camera tokens and an \textbf{Asymmetric Dual-Flow Decoder (ADF-Decoder)} that enforces directionally constrained communication between camera and image tokens. This maintains clean factorization within a feed-forward architecture, enabling coherent reconstruction and stable pose estimation without iterative refinement.
Extensive experiments demonstrate that TokenSplat achieves higher reconstruction fidelity and novel-view synthesis quality in pose-free settings, and significantly improves pose estimation accuracy compared to prior pose-free methods.  Project page: \textcolor{blue}{https://kidleyh.github.io/tokensplat/}\href{https://kidleyh.github.io/tokensplat/}.
\end{abstract}    
\section{Introduction}
\label{sec:intro}

3D Gaussian Splatting (3DGS) \cite{kerbl3Dgaussians} has recently emerged as an efficient alternative to neural radiance fields (NeRFs) \cite{mildenhall2021nerf}, enabling high-quality rendering with significantly faster performance. Despite this progress, most existing 3DGS-based reconstruction pipelines \cite{yu2024mip, lu2024scaffold, li2025micro, wang2021nerf-, lin2021barf, chng2022gaussianpose} rely on per-scene optimization, which restricts their scalability and limits generalization across diverse scenes. Recent feed-forward variants \cite{charatan2024pixelsplat, chen2024mvsplat, wewer2024latentsplat, zou2024triplane, keetha2025mapanything} alleviate this by predicting 3D Gaussians directly from input images, but their applicability remains constrained by the requirement for accurate camera poses. Pose estimation via structure-from-motion (SfM) \cite{schonberger2016structure} is computationally expensive and prone to failure in challenging environments, significantly impacting reconstruction stability.

%Recently, some methods \cite{chen2023dbarf, smith2023flowcam, hong2024unifying} have integrated pose estimation and scene reconstruction into a unified pipeline through alternating process, but their performance still lags significantly behind pose-required approaches \cite{chen2024mvsplat}. More recently, several pose-free methods \cite{jiang2023leap, smart2024splatt3r, wang2023pf, ye2024nopo} attempt to directly learn 3DGS scene representations and estimate the corresponding camera poses from sparse, pose-unknown images in a canonical space. However, these methods predict both the 3D scene and camera pose from the same image features. This use of a single feature representation fails to effectively disentangle camera information from other feature content, which severely limits the accuracy of camera pose prediction. On the other hand, existing methods use a pixel-aligned 3DGS head for 3DGS prediction, which is only suitable for reconstruction from a small number of sparse images. As the number of images increases and becomes denser, generating the same number of 3DGS as pixels leads to significant redundancy in 3D space, causing a substantial decline in scene quality. Some methods \cite{wang2024freesplat, jiang2025anysplat} attempt to directly fuse 3DGS attributes to address the redundancy problem in denser image settings. However, such 3D GS aggregation methods are designed only at the local level, with limited fusion receptive fields, which often leads to issues such as scene fragmentation, floating artifacts, and visual artifacts.

To reduce dependence on external poses, several approaches \cite{chen2023dbarf, smith2023flowcam, hong2024unifying} jointly optimize camera poses and reconstruction through alternating pipelines. While effective in alleviating strict pose requirements, these methods still struggle to maintain reconstruction fidelity when many or densely captured images are involved. More recent pose-free frameworks \cite{jiang2023leap, smart2024splatt3r, wang2023pf, ye2024nopo} attempt to infer both camera poses and 3D structure directly from sparse unposed images. However, these approaches typically encode scene information and viewpoint cues within the same feature embeddings, leading to representation entanglement that camera parameters become difficult to disentangle from scene content, and errors in pose estimation propagate to 3D reconstruction.

Furthermore, most existing pose-free methods rely on pixel-aligned 3DGS heads that generate Gaussians at pixel-level granularity. As the number of views increases, this pixel-centric prediction induces redundant and overlapping Gaussians, causing geometric blurring, color inconsistency, and reduced scene fidelity. Although some works \cite{wang2024freesplat, jiang2025anysplat} attempt multi-view 3DGS attribute fusion to suppress redundancy, their local aggregation scope limits their ability to form globally coherent structure, often resulting in fragmented or unstable reconstructions.

To address these challenges, we propose TokenSplat, a feed-forward 3D Gaussian splatting framework that reconstructs 3D scenes from an arbitrary number of unposed images while jointly estimating camera poses.
At its core lies the \textit{Token-aligned Gaussian Prediction} module, which aligns semantically corresponding information across viewpoints directly in the feature space.
Guided by coarse token positions and fusion confidence, this module adaptively aggregates multi-scale contextual features across views, enabling long-range cross-view reasoning and reducing redundancy from overlapping Gaussians.
The aggregated tokens are subsequently decoded by a 3D Gaussian prediction head, which maps each token to multiple Gaussians.
This one-to-many mapping decouples Gaussian density from pixel resolution, yielding denser and more expressive splats while preserving both structural integrity and semantic coherence.

To jointly optimize 3D reconstruction and camera pose estimation within a feed-forward architecture, we introduce learnable camera tokens and an \textit{Asymmetric Dual-Flow Decoder (ADF-Decoder)} that regulates the interaction between pose reasoning and scene reconstruction. Without iterative refinement loops traditionally used to disentangle pose and geometry, maintaining clean factorization between camera parameters and scene features becomes crucial. Unlike symmetric attention mechanisms that allow bidirectional mixing and risk entangling viewpoint-specific cues with scene semantics, the ADF-Decoder imposes directionally constrained information flow between camera tokens and image tokens. Camera tokens extract robust geometric cues from image tokens to support accurate pose prediction, while only stabilized low-frequency pose alignment signals are propagated back to image tokens. Through disentangled yet mutually supportive representations, TokenSplat achieves stronger cross-view alignment, more stable pose estimation, and higher-fidelity reconstruction.

In summary, our main contributions are as follows:

\begin{itemize}
\item We propose \textit{TokenSplat}, a feed-forward pose-free reconstruction framework that jointly estimates camera poses and 3D Gaussian scenes from unposed multi-view images, exhibiting strong generalization.
\item We introduce \textit{Token-aligned Gaussian Prediction} module that enables relatively long-range multi-view feature aggregation and high-quality Gaussian generation with reduced redundancy and fragmentation.
\item We design an \textit{Asymmetric Dual-Flow Decoder} that disentangles pose reasoning from scene encoding while enabling mutual reinforcement, leading to more accurate pose estimation and improved reconstruction fidelity.
\end{itemize}

\section{Related Work}

\subsection{Feed-forward 3D Reconstruction}

Neural radiance fields (NeRF) \cite{mildenhall2021nerf} and 3D Gaussian Splatting (3DGS) \cite{kerbl3Dgaussians} have achieved impressive results in novel view synthesis. However, their scene-specific optimization restricts scalability and generalization across multiple scenes. To overcome this limitation, recent works \cite{charatan2024pixelsplat, chen2024mvsplat, zhang2024gslrm, tang2024lgm, tang2024hisplat, xu2024grm, chen2021mvsnerf, xu2024murf} have explored feed-forward approaches. These methods can reconstruct scenes without per-scene optimization and generalize to unseen environments. Some works \cite{wang2024freesplat, xu2025depthsplat} further improve model efficiency and capability by incorporating lightweight cost volume modules or introducing depth priors. However, most of these methods are still only applicable under sparse-view conditions, and their broader applicability remains limited due to the requirement for precise image poses as input.

\subsection{Pose-Free 3D Scene Reconstruction}

Feed-forward methods that rely on camera parameters \cite{chen2024mvsplat, zhang2025pansplat, wang2025volsplat} typically obtain them via structure-from-motion (SfM) tools such as COLMAP \cite{schonberger2016structure}, a process that can be complex and error-prone. Some approaches jointly optimize camera poses and scene representations, but they often require inefficient pipelines. For instance, PF3Splat \cite{hong2024pf3plat} relies on off-the-shelf feature descriptors (LightGlue \cite{lindenberger2023lightglue}) and RANSAC-based \cite{fischler1981random} pose initialization, and SelfSplat uses cross-view U-Nets \cite{ronneberger2015u} to predict depth and pose separately. More recently, pose-free methods \cite{jiang2023leap, wang2023pf, ye2024nopo, huang2025spf} reconstruct scenes directly in a canonical space without explicit pose inputs, leveraging point or feature matching. While these methods improve robustness with sparse inputs, they still rely on primitive pixel-aligned 3D Gaussians, which limits multi-view reconstruction fidelity and pose estimation accuracy.

To reduce redundancy and improve reconstruction consistency, FreeSplat \cite{wang2024freesplat} maintains global Gaussians and continuously fuses new view-specific Gaussians, while AnySplat \cite{jiang2025anysplat} follows the VGGT \cite{wang2025vggt} paradigm and directly performs voxel fusion on the predicted Gaussians. However, their purely local fusion of 3DGS attributes in 3D space lacks broader contextual reasoning, often resulting in fragmented or inconsistent scenes. In contrast, our method performs multi-view feature aggregation at the token level, integrating semantically corresponding features across views based on coarse token positions and confidence scores. This alleviates redundancy in overlapping regions, and produces structurally coherent and complete reconstructions.

\begin{figure*}
    \centering
    \includegraphics[width=0.98\textwidth]{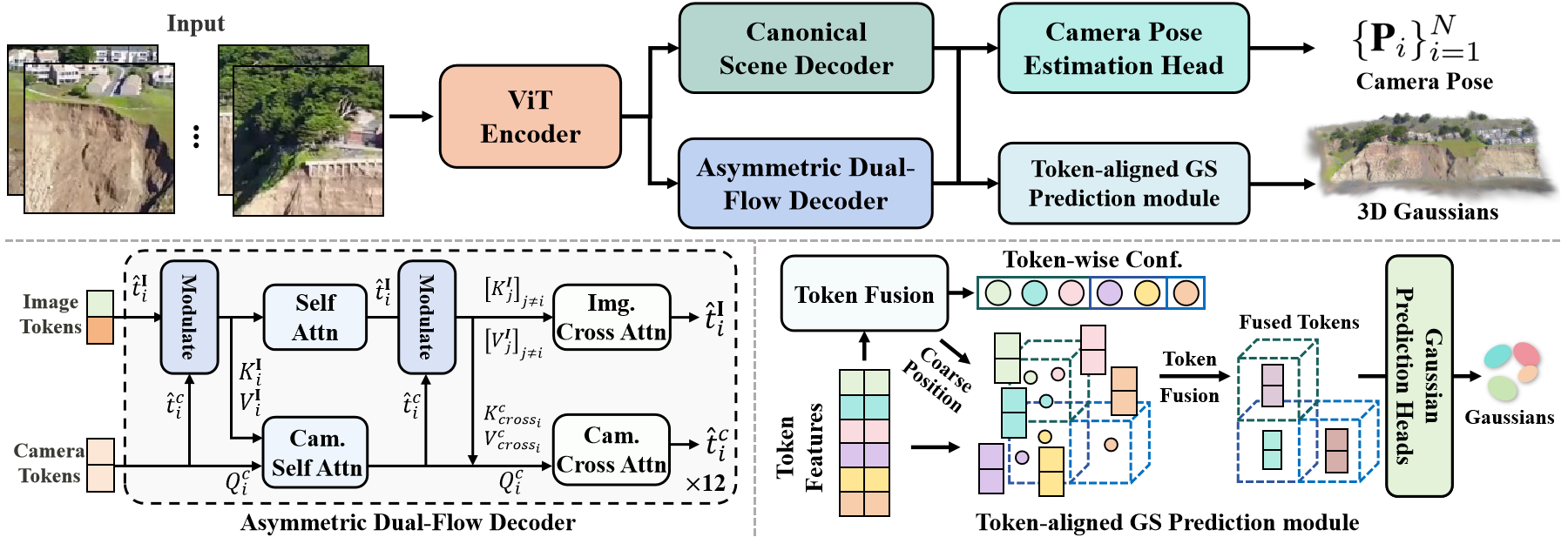}
    \caption{Overview of TokenSplat. TokenSplat performs feed-forward 3D Gaussian reconstruction and camera pose estimation from unposed images. A shared ViT encoder extracts image tokens, which are processed by the Canonical Scene Decoder and the Asymmetric Dual-Flow Decoder (ADF-Decoder). The fused tokens are then used by the Token-aligned Gaussian Prediction module and the camera pose head to generate dense 3D Gaussians and accurate poses.}
    \label{fig:pipeline}
\end{figure*}

\section{Method}
% Specifically, \(\textbf{P} = [\textbf{R} | \textbf{T}]\), where \(\textbf{R} \in \mathbb{R}^{3 \times 3}\) represents the rotation matrix, and \(\textbf{T} \in \mathbb{R}^{3 \times 1}\) represents the translation vector.
\subsection{Problem Formulation}

We aim to learn a feed-forward network that jointly reconstructs 3D Gaussians and predicts camera poses from a sequence of $N$ unposed images $\{\textbf{I}_i\}_{i=1}^N$, where $\textbf{I}_i\in \mathbb{R}^{H\times W\times 3}$. For 3D reconstruction, we predict a set of 3D Gaussians in a canonical 3D space:
\begin{equation}
\left\{ (\mu_g, \sigma_g, r_g, s_g, c_g) \right\}_{g=1}^G,
\end{equation}
where each Gaussian is parameterized by its center $\mu_g \in \mathbb{R}^3$, opacity $\sigma_g \in \mathbb{R}^+$, rotation quaternion $r_g \in \mathbb{R}^4$, scale $s_g \in \mathbb{R}^3$, and spherical harmonics (SH) coefficients $c_g \in \mathbb{R}^{3\times(k+1)^2}$ of degree $k$, following \cite{kerbl20233d}. 
For camera pose estimation, the network predicts per-view poses $\mathbf{P}_i$ that transform each input image $\mathbf{I}_i$ into the canonical reference view $\mathbf{I}_1$.
Formally, the model learns a mapping:
\begin{equation}
f_\theta : \{ \textbf{I}_i \}_{i=1}^N \longmapsto \left\{ (\mu_g, \sigma_g, r_g, s_g, c_g) \right\}_{g=1}^G \cup \{ \textbf{P}_i \}_{i=1}^N.
\end{equation}

During training, both the 3D Gaussian attributes and camera poses are jointly optimized. %During evaluation, we focus on two downstream tasks: novel view synthesis and multi-view camera pose estimation.

\subsection{Architecture}

As illustrated in Fig.~\ref{fig:pipeline}, our method is an Transformer-based architecture for feed-forward 3D reconstruction from unposed images. Each input view is first encoded into image tokens via a shared \textbf{\textit{ViT Encoder}}. The reference view $\mathbf{I}_1$ is processed by a \textbf{\textit{Canonical Scene Decoder}}, which uses cross-attention to integrate information from other views and establish a canonical scene representation. The remaining views are processed by the \textbf{\textit{Asymmetric Dual-Flow Decoder (ADF-Decoder)}}, which refines both image and learnable camera tokens through directionally constrained communication, effectively disentangling camera parameters from image features. The outputs of these decoders are then utilized in two parallel branches: the \textbf{\textit{Camera Pose Estimation Head}} predicts per-view camera transformations, while the \textbf{\textit{Token-aligned Gaussian Prediction module}} aggregates multi-view tokens in the feature space and generates dense, coherent 3D Gaussians through its prediction head.
This unified feed-forward design enables consistent multi-view reconstruction and robust pose estimation without any iterative refinement.

%As illustrated in Fig.~\ref{fig:pipeline}, our method is an end-to-end Transformer-based architecture for feed-forward 3D reconstruction from unposed images. Each input view is first encoded into image tokens via a shared \textbf{ViT Encoder}. The reference view $\textbf{I}_1$ is processed by a \textbf{multi-view ViT decoder} with cross-attention to aggregate information across all views, establishing a canonical scene representation. The remaining views are handled by the \textbf{Asymmetric Dual-Flow Decoder (ADF-Decoder)}, which refines image and learnable camera tokens through directionally constrained attention, effectively disentangling camera parameters and scene features. The outputs of both decoders are then used in parallel: the \textbf{pose head} predicts per-view camera transformations, while \textbf{TokenFusion} aggregates multi-view features to feed the token-aligned Gaussian prediction head, generating dense 3D Gaussians. This design enables coherent multi-view reconstruction and robust pose estimation in a single feed-forward pass without iterative refinement.

\noindent\textbf{ViT Encoder.}  
For each input view, the RGB image is divided into patches and embedded as a sequence of image tokens. Optionally, following \cite{ye2024nopo}, camera intrinsics are encoded as an additional token via a linear layer and concatenated with the image tokens along the spatial dimension to mitigate scale ambiguity. Each view's tokens are independently processed by a weight-shared ViT encoder, without cross-view interaction.

%\noindent\textbf{Multi-view ViT Decoder.}  
%Following \cite{ye2024nopo, huang2025spf}, the model adopts the camera viewpoint of \(\textbf{I}_1\) as the canonical space and only needs to predict the camera transformations of the remaining views relative to \(\textbf{I}_1\).
%The reference view $\textbf{I}_1$ is decoded using a ViT decoder with cross-attention to integrate features from all views, forming a canonical scene representation. This design ensures that multi-view context is captured efficiently for the reference view.

\noindent\textbf{Canonical Scene Decoder.}
To establish a canonical scene representation, the reference view $\mathbf{I}_1$ is decoded using a ViT decoder with cross-attention to other views. This ensures that multi-view context is efficiently captured for the reference view and provides a consistent basis for reconstructing the 3D scene. The canonical representation facilitates the subsequent refinement of camera and scene features for the remaining views.

\subsection{Asymmetric Dual-Flow Decoder}

The ADF-Decoder is designed to jointly refine image and camera representations while explicitly disentangling scene content from camera parameters. To prevent entangling viewpoint-specific cues with scene semantics, it employs an \textit{asymmetric update scheme} that image tokens primarily aggregate scene context, whereas camera tokens extract geometric cues from image tokens and propagate only stabilized pose signals back. This design enables accurate pose estimation and high-fidelity 3D reconstruction in a single feed-forward pass.

\noindent\textbf{Token Initialization.}  
Image tokens are generated by the shared ViT encoder. Camera tokens are initialized from learnable embeddings and duplicated for the $N-1$ non-reference views to provide per-view pose representations. At each decoder layer, camera tokens interact with the corresponding image tokens through directionally constrained attention, allowing them to extract view-specific geometric cues while preventing unnecessary feature mixing. 

\noindent\textbf{Self-Attention.}  
The ADF-Decoder consists of 12 decoder blocks. Within each block, self-attention is applied separately to image and camera tokens. Image tokens integrate intra-view context to capture local scene structures and enhance reconstruction. Camera tokens attend to their corresponding image tokens to extract geometric cues for pose estimation. Formally, for view $i$, let $\hat{t}^\mathbf{I}_i$ denote the image tokens and $\hat{t}^c_i$ denote the camera token. They are projected to queries, keys, and values $Q^\mathbf{I}_i$ ($Q^c_i$), $K^\mathbf{I}_i$, and $V^\mathbf{I}_i$.
%\[
%Q^\mathbf{I}_i, K^\mathbf{I}_i, V^\mathbf{I}_i = \text{Linear}(\hat{t}^\mathbf{I}_i), \quad
%Q^c_i = \text{Linear}(\hat{t}^c_i).
%\]
The updates are computed as:
\begin{align}
\hat{t}^\mathbf{I}_i &\leftarrow \text{Softmax}\Big(Q^\mathbf{I}_i {K^\mathbf{I}_i}^\top / \sqrt{d}\Big) V^\mathbf{I}_i,\\
\hat{t}^c_i &\leftarrow \text{Softmax}\Big(Q^c_i {K^\mathbf{I}_i}^\top / \sqrt{d}\Big) V^\mathbf{I}_i,
\end{align}
where $d$ is the corresponding dimensionality of the queries and keys. By updating camera tokens through their interaction with image tokens of the same view, the decoder extracts robust geometric features to support pose prediction.

\noindent\textbf{Image Token Cross-View Attention.}  
To capture global multi-view context and enforce inter-view consistency, each view performs cross-attention over tokens from other views while explicitly avoiding interactions with its own tokens. This prevents information leakage across views and ensures complementary cues are integrated from the rest of the scene. For the image tokens of view $\mathbf{I}_i$, the keys and values from other views are concatenated along the spatial dimension:
\begin{align}
    [K^\mathbf{I}_j]_{j\ne i} &= K^\mathbf{I}_1 || \cdots || K^\mathbf{I}_{i-1} || K^\mathbf{I}_{i+1} || \cdots || K^\mathbf{I}_N, \\
    [V^\mathbf{I}_j]_{j\ne i} &= V^\mathbf{I}_1 || \cdots || V^\mathbf{I}_{i-1} || V^\mathbf{I}_{i+1} || \cdots || V^\mathbf{I}_N.
\end{align}
The image tokens are then updated via cross-attention:
\begin{equation}
    \hat{t}^\mathbf{I}_i \leftarrow \text{Softmax}\Big(Q^\mathbf{I}_i {[K^\mathbf{I}_j]_{j\ne i}}^\top / \sqrt{d}\Big) [V^\mathbf{I}_j]_{j\ne i}.
\end{equation}
To reduce computation with many views, a hyperparameter $pnv$ limits each view's attention to its $pnv-1$ neighbors, balancing context and efficiency.

\noindent\textbf{Camera Token Cross-Attention.}  
% Camera tokens encode per-view pose information.
Updating camera tokens solely via other camera tokens or solely via image tokens is insufficient to capture global geometry. Therefore, each camera token interacts with both image and camera tokens from other views, allowing it to aggregate multi-view geometric cues for more accurate camera pose estimation.

For view $\mathbf{I}_i$, let $\hat{t}^c_i$ be its camera token. All camera tokens are projected to obtain $K^c$ and $V^c$. Each $\hat{t}^c_j$ for $j \ne i$ is replicated to match the number of image tokens of other views, and then summed with the image token keys and values to form the cross-attention inputs:
\begin{align}
    K^c_{cross_i} &= [K^\mathbf{I}_j]_{j\ne i} + [\text{repeat}(K^c_j)]_{j \ne i},\\
    V^c_{cross_i} &= [V^\mathbf{I}_j]_{j\ne i} + [\text{repeat}(V^c_j)]_{j \ne i}.
\end{align}
The camera token is updated as:
\begin{equation}
    \hat{t}^c_i \leftarrow \text{Softmax}\Big(Q^c_i {K^c_{cross_i}}^\top / \sqrt{d}\Big) V^c_{cross_i}.
\end{equation}

\noindent\textbf{Pre- and Post-Modulation.}  
Due to differences in token counts and information content between image and camera tokens, image tokens are modulated with corresponding camera tokens both before and after attention. This stabilizes updates, reinforces disentanglement between pose and scene features, and ensures that pose cues guide reconstruction without contaminating scene semantics.

%\subsection{Token-aligned Gaussian Prediction}
\subsection{Token Fusion for Scene Reconstruction}

The \textbf{Token-aligned Gaussian Prediction} module aggregates multi-view features and predicts dense 3D Gaussians for scene reconstruction. It comprises a \textbf{\textit{Token Fusion}} operation and a \textbf{\textit{Gaussian Prediction Head}}. Unlike traditional pixel-aligned Gaussian methods, which directly predict dense Gaussians for each pixel and suffer from redundancy and geometric blurring in dense multi-view setups, our method fuses features across tokens to produce more consistent and efficient 3D Gaussians.

%The TokenFusion module consists of token fusion operation and Gaussian prediction heads. Unlike traditional pixel-aligned Gaussian heads, we first fuse token features from different viewpoints in the feature space. Therefore, we use token fusion operantion to predict the coarse positions and fusion confidence of the tokens. The tokens are then grouped based on their coarse positions and fused according to their confidence scores. For the fusion tokens, the Gaussian Prediction heads predict the positional offsets and Gaussian parameters of each Gaussian relative to its corresponding token. The token fusion part adopts the DPT-based architecture \cite{ranftl2021vision}, and Gaussian prediction heads is a novel architecture composed of MLPs.

\noindent\textbf{Token Fusion.}  
The token fusion operation first predicts coarse positions and fusion confidence for each token. Tokens are then grouped based on spatial proximity with a spatial grouping size $\epsilon$, and fused using softmax-normalized confidence scores. This produces a set of merged tokens with consolidated features and coarse positions. The fusion network adopts a DPT-based architecture \cite{ranftl2021vision}, similar to \cite{ye2024nopo}, with the number of predicted parameters adjusted.

%First, we use the \textbf{token fusion operation} to predict the coarse positions and fusion confidence of the tokens. We simply group the tokens according to their coarse positions with a grouping size of $\epsilon$, and perform feature fusion using the softmax-normalized confidence scores to obtain the coarse positions and fused features of the merged tokens. The token fusion head also adopts a DPT architecture, similar to that in \cite{ye2024nopo}, with the only difference being the number of predicted parameters.

\noindent\textbf{Gaussian Prediction Head.}  
For each fused token, the Gaussian prediction heads generate denser Gaussians, including positional offsets relative to the token and Gaussian attributes. Multi-scale features \(\{F_i\}_{i=1}^{n_l}\) from different layers of the Transformer decoder corresponding to the fused tokens are first upsampled and linearly projected:
\begin{equation}
    \hat{F}_i = Proj_i(F_i), \quad i = 1,\dots,n_l.
\end{equation}
Channel dimensions are unified, and the features are upsampled according to layer scale (larger upsampling for lower layers, smaller for higher layers). Features are then progressively fused from deep to shallow layers using the residual fusion module \(RF\) composed of residual blocks and upsampling:
\begin{align}
    F^{fusion}_{n_l} &= RF_{n_l}(\hat{F}_{n_l}),\\
    F^{fusion}_i &= RF_i(\hat{F}_i, F^{fusion}_{i+1}).
\end{align}
Residual blocks enhance feature expressiveness, while upsampling spatially aligns semantic and detailed information. The resulting fused features combine fine-grained details with rich semantic context.

Finally, the fused high-resolution features are passed through a prediction block consisting of upsampling and linear layers, producing continuous tensors representing the final Gaussian predictions. For full architectural details of the token-aligned Gaussian prediction heads, please refer to the appendix.

\subsection{Camera Pose Estimation Head}  
The camera pose estimation head predicts the extrinsic parameters for each input view (excluding the reference view \(\textbf{I}_1\)). It takes the per-view camera tokens produced by the decoder as input and applies a linear projection to regress the camera transformation \(\mathbf{P}\) for each view. Jointly optimizing camera poses alongside the token-aligned Gaussian reconstruction encourages the network to learn geometrically consistent 3D features, which improves both pose estimation and scene reconstruction accuracy.

\subsection{Loss Functions}
\noindent\textbf{Image Rendering Loss.}  
We supervise the model using the ground-truth target RGB images. The rendering loss combines an L2 loss and a perceptual LPIPS loss \cite{zhang2018unreasonable}:
\begin{equation}
    \mathcal{L}_{render} = \mathcal{L}_2(\mathbf{I}, \hat{\mathbf{I}}) + \lambda_{lpips} \mathcal{L}_{lpips}(\mathbf{I}, \hat{\mathbf{I}}),
\end{equation}
where $\mathbf{I}$ and $\hat{\mathbf{I}}$ denote the ground-truth and rendered images, and $\lambda_{lpips}$ balances the perceptual term.

\noindent\textbf{Camera Pose Loss.}  
The camera pose head predicts each view's extrinsic parameters relative to the reference view $\mathbf{I}_1$. We supervise the predictions with a combination of MSE and Unit Dual Quaternion (DQ) alignment loss \cite{clifford1871preliminary,daniilidis1999hand,kavan2006dual}, which jointly represents rotation and translation to avoid inconsistencies from separate predictions. The alignment loss is defined as:
\begin{equation}
    \mathcal{L}_{align} = \| p^I - p \hat{p}^{\ast} \| + \| p^I - \hat{p} p^{\ast} \|,
\end{equation}
where $p$ and $\hat{p}$ are the predicted and ground-truth unit DQs, and $p^{\ast}$ denotes the conjugate. The overall camera pose loss is:
\begin{equation}
    \mathcal{L}_{pose} = \mathcal{L}_{MSE}(\mathbf{P}, \hat{\mathbf{P}}) + \mathcal{L}_{align}.
\end{equation}
%where $\textbf{P}$ and $\hat{\textbf{P}}$ denote the ground-truth and predict pose, respectively.

\noindent\textbf{Total Loss.}  
The model is trained end-to-end with the total loss:
\begin{equation}
    \mathcal{L} = \mathcal{L}_{render} + \lambda_c \mathcal{L}_{pose},
\end{equation}
where $\lambda_c$ balances the contribution of the camera pose supervision.

\begin{table*}[t!]
\renewcommand\arraystretch{0.75} 
\setlength{\tabcolsep}{1.5pt}
\centering

\caption{Quantitative results of NVS on RE10K with varying reference views (left) and cross-dataset generalization to ScanNet (right). The \textbf{best} and \underline{second-best} values are highlighted.
}
% \resizebox{0.98\columnwidth}{!}{
\begin{tabular}{llcccccc|cccccc}
\toprule
& \multirow{2}{*}{Method} & \multicolumn{3}{c}{RE10K (4 views)} & \multicolumn{3}{c|}{RE10K (8 views)} & \multicolumn{3}{c}{ScanNet (4 views)} & \multicolumn{3}{c}{ScanNet (8 views)}\\ 
\cmidrule(lr){3-5} \cmidrule(lr){6-8} \cmidrule(lr){9-11} \cmidrule(lr){12-14}
&  & PSNR$\uparrow$ & SSIM$\uparrow$ & LPIPS$\downarrow$ &  PSNR$\uparrow$ & SSIM$\uparrow$ & LPIPS$\downarrow$ & PSNR$\uparrow$ & SSIM$\uparrow$ & LPIPS$\downarrow$ &  PSNR$\uparrow$ & SSIM$\uparrow$ & LPIPS$\downarrow$\\
\midrule
\multirow{2}{*}{\makecell[l]{Pose-\\[-1ex]required}} 
& MVSplat & 23.82 & 0.792 & 0.201 & 23.96 & 0.802 & 0.164 & 24.48 & 0.854 & 0.225 & 23.11 & 0.761 & 0.296 \\
& FreeSplat & \underline{24.99} & 0.814 & \underline{0.162} & \underline{25.20} & \underline{0.829} & 0.179  & \underline{27.23} & \underline{0.873} & 0.196 & \underline{24.64} & \underline{0.819} & \underline{0.241} \\
\midrule
\multirow{7}{*}{\makecell[l]{Pose-\\[-1ex]free}} 
& NoPoSplat & 24.87 & 0.813 & 0.169 & 25.01 & 0.832 & \underline{0.163} & 27.10 & 0.864 & 0.189 & 24.26 & 0.796 & 0.269\\
& VicaSplat & 24.65 & 0.795 & 0.206 & 24.50 &  0.806  & 0.164 & 26.46 & 0.837 & 0.214 & 22.95 & 0.765 & 0.306 \\
& SPFSplat & 24.98 & \underline{0.819} & 0.166 & 25.18 & 0.828 & 0.169 & 27.17 & 0.873 & \underline{0.185} & 24.15 & 0.793 & 0.263 \\
& Anysplat$\ast$ & 20.28 & 0.657 & 0.251 & 20.93 & 0.746 & 0.284 & 22.74 & 0.768 & 0.242 & 21.83 & 0.673 & 0.358 \\
& Anysplat & 15.55 & 0.475 & 0.424 & 16.23 & 0.520 & 0.401 & 17.07 & 0.580 & 0.436 & 16.72 & 0.577 & 0.457 \\
% & \textbf{Ours-PGH} & & & &  \\
& \textbf{Ours} & \textbf{25.14} & \textbf{0.834} & \textbf{0.156} & \textbf{26.15} & \textbf{0.858} & \textbf{0.135}  & \textbf{28.15} & \textbf{0.884} & \textbf{0.178} & \textbf{25.15} & \textbf{0.824} & \textbf{0.233} \\
\bottomrule
\end{tabular}
\label{tab:re10k-nvs}
\end{table*}

\begin{table*}[t!]
\renewcommand\arraystretch{0.75}
\centering
\caption{Quantitative results of NVS on ScanNet under varying numbers of views. The \textbf{best} and \underline{second-best} values are highlighted.}
\begin{tabular}{llccccccccc}
\toprule
& \multirow{2}{*}{Method} & \multicolumn{3}{c}{3 views} & \multicolumn{3}{c}{10 views} & \multicolumn{3}{c}{28 views} \\ 
\cmidrule(lr){3-5} \cmidrule(lr){6-8} \cmidrule(lr){9-11} 
&  & PSNR$\uparrow$ & SSIM$\uparrow$ & LPIPS$\downarrow$ & PSNR$\uparrow$ & SSIM$\uparrow$ & LPIPS$\downarrow$ & PSNR$\uparrow$ & SSIM$\uparrow$ & LPIPS$\downarrow$ \\
\midrule
\multirow{2}{*}{\makecell[l]{Pose-\\[-1ex]required}} & MVSplat & 24.98 & 0.784 & 0.228 & 23.12 & 0.768 & 0.236 & 22.26 & 0.714 & 0.336 \\
& FreeSplat & 25.63 & \underline{0.804} & 0.218 & \underline{25.38} & \underline{0.815} & \underline{0.189} & \underline{24.30} & \underline{0.798} & \underline{0.249} \\
\midrule
\multirow{7}{*}{\makecell[l]{Pose-\\[-1ex]free}} 
& NoPoSplat & 25.55 & 0.803 & \underline{0.213} & 24.21 & 0.811 & 0.198 & 23.21 & 0.764 & 0.268 \\
& VicaSplat & 24.54 & 0.786 & 0.226 & 23.73 & 0.760 & 0.268 & 22.37 & 0.709 & 0.316 \\
& SPFSplat & \underline{25.85} & 0.788 & 0.229 & 24.63 & 0.773 & 0.226 & 23.54 & 0.775 &	0.261 \\
& Anysplat$\ast$ & 19.84 & 0.658 & 0.314 & 22.06 & 0.687 & 0.353 & 21.17 & 0.680 & 0.357 \\
& Anysplat & 15.05 & 0.415 & 0.416 & 16.73 & 0.522 & 0.437 & 16.58 & 0.561 & 0.439 \\
% & \textbf{Ours-PGH} & \textbf{26.55} & \textbf{0.828} & \textbf{0.183} & 25.19 & 0.829 & 0.212 & 22.72 & 0.780 & 0.265 \\
& \textbf{Ours} & \textbf{26.57} & \textbf{0.841} & \textbf{0.189} & \textbf{26.83} & \textbf{0.851} & \textbf{0.179} & \textbf{26.87} & \textbf{0.859} & \textbf{0.173} \\
\bottomrule
\end{tabular}
\label{tab:scannet-nvs}
\end{table*}

% \begin{figure*}[t!]
% \centering
% \includegraphics[width=0.96\textwidth]{img/re10k_nvs_less.png}
% \caption{Qualitative comparison on RE10K under varying numbers of reference views.}
% \label{fig:re10k_nvs}
% \end{figure*}

\section{Experiment}
%We report evaluation results on novel view synthesis quality for both sparse and long image sequences on real-world datasets, as well as pose estimation results on several datasets.
We evaluate our method on novel view synthesis (NVS) and camera pose estimation across sparse and long-sequence real-world datasets.

\subsection{Experimental Settings}
\noindent\textbf{Datasets.}
%To evaluate novel view synthesis (NVS) and camera pose estimation on different datasets, we conducted experiments on the real-world datasets ScanNet \cite{dai2017scannet} and RealEstate10K (RE10K) \cite{zhou2018stereo}. For ScanNet, we followed the training and testing scene splits as in \cite{wang2024freesplat} for both training and evaluation. Similarly, for the RE10K dataset, we adopted the splits as in \cite{ye2024nopo} for training and testing.
We conduct experiments on ScanNet \cite{dai2017scannet} and RealEstate10K (RE10K) \cite{zhou2018stereo}.
For ScanNet, we adopt the training/testing splits from \cite{wang2024freesplat}, and for RE10K we follow \cite{ye2024nopo}. Both settings ensure a direct and consistent comparison with prior work.

\noindent\textbf{Baselines and Metrics.}
We evaluate NVS performance using PSNR, SSIM \cite{wang2004image}, and LPIPS \cite{zhang2018unreasonable}.
For camera pose estimation, we report Absolute Translation Error (ATE), Relative Translation Error (RPE-t), and Relative Rotation Error (RPE-r).
We compare against generalizable NVS methods, including
%We evaluate NVS with common photo metrics: PSNR, SSIM \cite{wang2004image} and LPIPS \cite{zhang2018unreasonable}. For camera pose estimation, we compute the Absolute Translation Error (ATE), Relative Translation Error (RPE-t), and Relative Rotation Error (RPE-r). We compare our method with state-of-the-art generalizable approaches for novel view synthesis. Specifically, we consider: 
(1) Pose-required: MVSplat \cite{chen2024mvsplat} and FreeSplat \cite{wang2024freesplat}; and (2) Pose-free: NoPoSplat \cite{ye2024nopo}, VicaSpalt \cite{li2025vicasplat}, SPFSplat \cite{huang2025spf}, and AnySplat \cite{jiang2025anysplat}.
Pose estimation comparisons are conducted only with pose-free methods.
%\textcolor{red}{Specifically, FreeSplat maintains global Gaussians and continuously fuses Gaussians from new views; VicaSplat enhances multi-view reconstruction by introducing cross-neighbor attention; and AnySplat directly fuses Gaussian attributes by predicting the fusion confidence of pixel-aligned Gaussians. These approaches respectively strengthen the ability to reconstruct from multiple views. Here, AnySplat refers to zero-shot results trained on other datasets, while AnySplat* denotes our results after fine-tuning on the corresponding dataset.}

\noindent\textbf{Implementation Details.}
Our method is implemented in PyTorch. The encoder follows \cite{ye2024nopo} and adopts a ViT-Large backbone with a patch size of 16. We initialize the encoder-decoder and Gaussian center head with MASt3R \cite{leroy2024grounding} weights, while the ADF-Decoder and remaining heads are randomly initialized.

Following \cite{li2025vicasplat}, we train and evaluate on RE10K under both 4-view and 8-view reference settings, and further perform cross-dataset generalization tests on ScanNet. For ScanNet, we adopt the 3-view and 10-view settings from \cite{wang2024freesplat}, where the 10-view configuration is obtained by augmenting the 3-view setup with additional reference views. To stress-test multi-view scalability on longer sequences, we also design a 28-view evaluation on ScanNet. 
% Since training under the 28-view setting is infeasible, all 28-view results are obtained using models trained with 10 views.

For fair comparisons, we follow \cite{chen2024mvsplat, li2025vicasplat, huang2025spf} and report all quantitative results at a fixed resolution of $256\times256$. Additional details are provided in the appendix.

\subsection{Experimental Results}
\noindent\textbf{Performance on Multi-view NVS.}
%As shown in Tab.~\ref{tab:re10k-nvs}, Tab.~\ref{tab:scannet-nvs}, Fig.~\ref{fig:re10k_nvs} and Fig.~\ref{fig:scannet_nvs}. Here, AnySplat refers to zero-shot results trained on other datasets, while AnySplat* denotes our results after fine-tuning on the corresponding dataset.
%As can be seen, the propsoed TokenSplat noticeably outperforms all state-of-the-art pose-free methods across datasets. It further demonstrates clear gains over multi-view approaches such as VicaSplat and AnySplat. VicaSplat enhances multi-view reconstruction by introducing cross-neighbor attention; and AnySplat directly fuses Gaussian attributes by predicting the fusion confidence of pixel-aligned Gaussians. These approaches respectively strengthen the ability to reconstruct from multiple views. while in dense-view reconstruction scenarios, TokenSplat even outperforms pose-dependent methods such as FreeSplat, surpassing it by 0.95 dB in PSNR on RE10K with 8 views.
The results are shown on the left side of Tab.~\ref{tab:re10k-nvs} and Tab.~\ref{tab:scannet-nvs}. Here, AnySplat refers to zero-shot results trained on other datasets, while AnySplat$\ast$ denotes the results we achieved after fine-tuning on the corresponding dataset. As can be seen, TokenSplat consistently outperforms state-of-the-art pose-free methods, including those specifically designed for multi-view input such as VicaSplat and AnySplat, which leverage cross-neighbor attention and pixel-aligned Gaussian fusion, respectively. In dense-view scenarios, TokenSplat even surpasses pose-dependent methods like FreeSplat, achieving 0.95 dB higher PSNR on RE10K with 8 views.

%For the 28-view evaluation, we use models trained under the 10-view setting. Even with this mismatch between training and evaluation view counts, our method maintains stable reconstruction quality, whereas competing methods show substantial degradation—including those specifically designed for multi-view aggregation. For example, our method outperforms AnySplat and FreeSplat by 0.168 and 0.091 in SSIM, respectively (AnySplat directly fuses Gaussian attributes by predicting the fusion confidence of pixel-aligned Gaussians). This difference stems from the fusion strategy: Gaussian-level fusion accumulates inconsistencies and redundancy as more views are added, degrading quality. In contrast, TokenSplat fuses in feature space, enhancing cross-view reasoning and reducing redundancy, especially in long-sequence, high-density scenarios.
For the 28-view evaluation, we use models trained with 10 views. Despite the difference in view counts, TokenSplat maintains stable reconstruction quality, while competing methods, including AnySplat, which fuses pixel-aligned Gaussians by predicting fusion confidence, and FreeSplat, show degradation. We attribute this to the Gaussian-level fusion strategy used in these methods, which accumulates inconsistencies and redundancies as more views are incorporated. In contrast, TokenSplat performs fusion in feature space, enhancing cross-view reasoning and reducing redundancy in long-sequence, high-density scenarios. Moreover, as the number of input images increases, our model achieves a higher SSIM of 0.061 over FreeSplat, while also showing improved novel view synthesis quality. These results  demonstrate the scalability of our approach to a larger number of views.

\begin{figure*}[t!]
\centering
\includegraphics[width=0.96\textwidth]{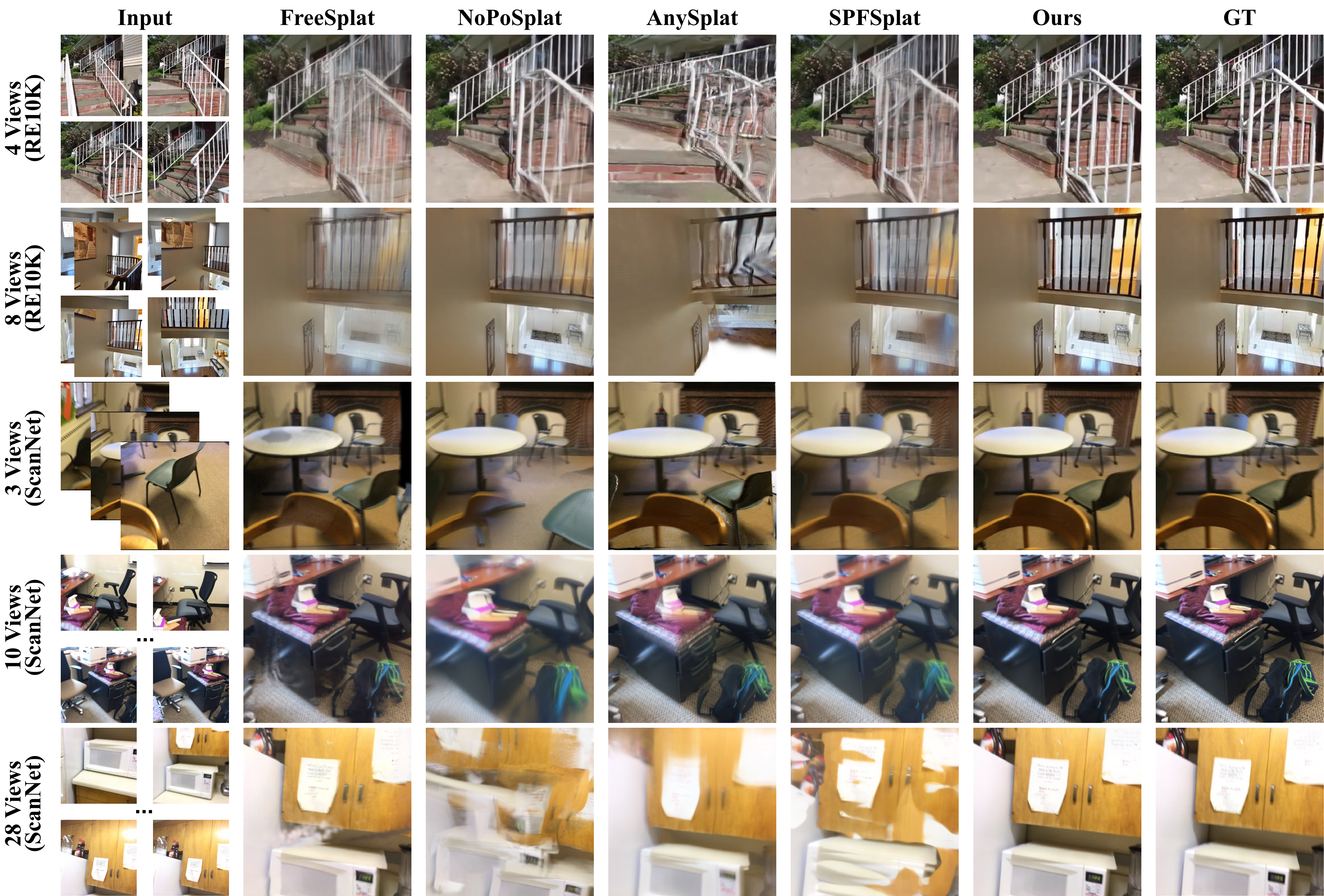}
\caption{Qualitative comparison on RE10K and ScanNet under varying numbers of reference views.}
\label{fig:qualitative_nvs_less}
\end{figure*}

In Fig.~\ref{fig:qualitative_nvs_less}, our method consistently produces clearer and more realistic novel view renderings in challenging cases. Ours better preserves fine structural details and reduces artifacts and blurring. See more visualizations in appendix.

\begin{figure*}[t!]
\centering
\includegraphics[width=0.96\textwidth]{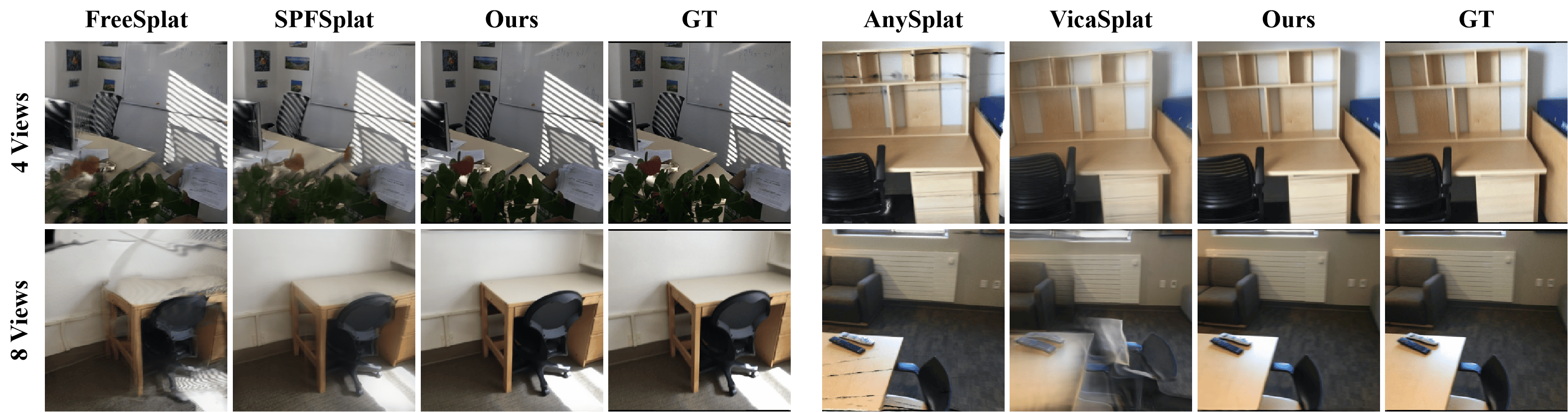}
\caption{Cross-dataset generalization from RE10K to ScanNet.}
\label{fig:re10k_cross_scannet_nvs}
\end{figure*}

\noindent\textbf{Pose Estimation.}
The results are shown in the left side of Tab.~\ref{tab:re10k-pose} and Tab.~\ref{tab:scannet-pose}. For NoPoSplat, poses are estimated via PnP \cite{hartley2003multiple} with RANSAC \cite{fischler1981random} from predicted Gaussian centers, while other methods directly output camera poses. TokenSplat consistently surpasses pose-free baselines on RE10K, reducing RPE-R (lower is better) by 0.335 and 0.147 compared to VicaSplat and AnySplat under the 8-view setting. This gain stems from our directionally constrained ADF-Decoder, which enforces disentangled interaction between camera and image tokens, leading to more stable pose learning. On ScanNet, the model maintains accurate pose estimation under the 28-view setting, reducing ATE by 0.018 over AnySplat, confirming both robustness and scalability of TokenSplat in multi-view scenarios.
%We evaluate Relative pose estimation between multi-view input images on RE10K and ScanNet, as shown in  Tab. \ref{tab:re10k-pose} and Tab. \ref{tab:scannet-pose}. For the NopoSplat method, pose estimation is performed using the predicted 3D Gaussian centers via PnP \cite{hartley2003multiple} with RANSAC \cite{fischler1981random}. For the other methods, we directly compute the results using their predicted camera poses. Our method consistently outperforms other pose-free approaches on the RE10K dataset. On RE10K with 8 views, our method further reduces RPE-R by 0.335 and 0.147 compared to VicaSplat and AnySplat, respectively. This improvement is attributed to our restricted communication strategy, which enables better disentanglement of camera tokens and leads to more accurate camera pose estimation, in contrast to their approaches where camera and image tokens freely exchange information in attention. Meanwhile, on the ScanNet dataset, even when extended to the 28-view setting, our model still maintains high novel view synthesis accuracy, \textcolor{red}{further reducing RPE-t by 0.113 compared to AnySplat}. This effectively demonstrates the accuracy and scalability of our model for multi-view pose prediction.

\begin{table*}[t!]
\renewcommand\arraystretch{0.75} 
\setlength{\tabcolsep}{3.5pt}
\centering
\caption{Quantitative results of pose prediction on RE10K with diverse views (left) and cross-dataset generalization to ScanNet (right).
%The \textbf{best} and \underline{second-best} values are highlighted.
}
\begin{tabular}{lcccccc|cccccc}
\toprule
\multirow{2}{*}{Method} & \multicolumn{3}{c}{RE10K (4 views)} & \multicolumn{3}{c|}{RE10K (8 views)} & \multicolumn{3}{c}{ScanNet (4 views)} & \multicolumn{3}{c}{ScanNet (8 views)}\\ 
\cmidrule(lr){2-4} \cmidrule(lr){5-7}  \cmidrule(lr){8-10} \cmidrule(lr){11-13} 
& ATE$\downarrow$ & RPE-t$\downarrow$ & RPE-r$\downarrow$ &  ATE$\downarrow$ & RPE-t$\downarrow$ & RPE-r$\downarrow$ & ATE$\downarrow$ & RPE-t$\downarrow$ & RPE-r$\downarrow$ &  ATE$\downarrow$ & RPE-t$\downarrow$ & RPE-r$\downarrow$ \\
\midrule
NoPoSplat &  0.032 & 0.070 & 1.923 & 0.123 & 0.135 & 1.757 & 0.139 & 0.259  & 2.135 & 0.208 & 0.268 & 4.098 \\
VicaSplat & \underline{0.027} & 0.057 & \underline{1.392} & 0.021 & 0.031 & 0.793 & 0.095 & 0.234 & 2.735 & 0.163 & 0.197 & 3.162 \\
SPFSplat & 0.036 & 0.078 & 3.113 & 0.037 & 0.047 & 2.117 & \underline{0.075} &  \underline{0.201} & 2.658 & 0.121 & 0.173 & 3.676 \\
Anysplat$\ast$ & 0.028 & \underline{0.051} & 1.515 & \underline{0.020} & \underline{0.025} &  \underline{0.578} & 0.092 & 0.238 & \underline{1.902} &  \underline{0.098} &  \underline{0.150} &  \underline{2.016} \\
Anysplat & 0.033 & 0.055 & 1.624 & 0.024 & 0.029 & 0.605 & 0.099 & 0.268 & 1.995 & 0.101 & 0.157 & 2.114\\
\textbf{Ours} & \textbf{0.016} & \textbf{0.034} & \textbf{1.054} & \textbf{0.012} & \textbf{0.019} & \textbf{0.458} & \textbf{0.062} & \textbf{0.159} & \textbf{1.162} & \textbf{0.088} & \textbf{0.130} & \textbf{1.672}\\
\bottomrule
\end{tabular}
\label{tab:re10k-pose}
\end{table*}

\begin{table*}[t!]
\renewcommand\arraystretch{0.75}
\setlength{\tabcolsep}{9.5pt}
\centering
\caption{Quantitative results of pose prediction on ScanNet with diverse views. The \textbf{best} and \underline{second-best} values are highlighted.}
\begin{tabular}{lccccccccc}
\toprule
\multirow{2}{*}{Method} & \multicolumn{3}{c}{3 views} & \multicolumn{3}{c}{10 views} & \multicolumn{3}{c}{28 views} \\ 
\cmidrule(lr){2-4} \cmidrule(lr){5-7} \cmidrule(lr){8-10} 
& ATE$\downarrow$ & RPE-t$\downarrow$ & RPE-r$\downarrow$ & ATE$\downarrow$ & RPE-t$\downarrow$ & RPE-r$\downarrow$ & ATE$\downarrow$ & RPE-t$\downarrow$ & RPE-r$\downarrow$ \\
\midrule
NoPoSplat &  \underline{0.055} & 0.218 & 2.041 & 0.167 & 0.175 & 5.537 & 0.187 & 0.197 & 4.551 \\
VicaSplat & 0.075  & 0.268 & 1.803 & 0.171 & 0.231 & 3.768 & 0.243 & 0.189 & 3.725  \\
SPFSplat & 0.104 & 0.336 & 2.351 & 0.155 &  \underline{0.167} & 3.249 & 0.207 & 0.216 & 4.428 \\
Anysplat$\ast$ & 0.057 & \underline{0.198} &  \underline{1.567} &  \underline{0.135} &  0.181 &  \underline{1.165} &  \underline{0.097} &  0.094 & \underline{1.239} \\
Anysplat & 0.062 & 0.213 & 1.611 & 0.148 & 0.196 & 1.192 & 0.098 & \underline{0.091} & 1.362 \\
\textbf{Ours} & \textbf{0.041} & \textbf{0.147} & \textbf{1.252} & \textbf{0.093} & \textbf{0.135} & \textbf{0.799} & \textbf{0.080} & \textbf{0.075} & \textbf{0.709} \\
\bottomrule
\end{tabular}
\label{tab:scannet-pose}
\end{table*}

\begin{figure*}[t!]
\centering
\includegraphics[width=0.93\textwidth]{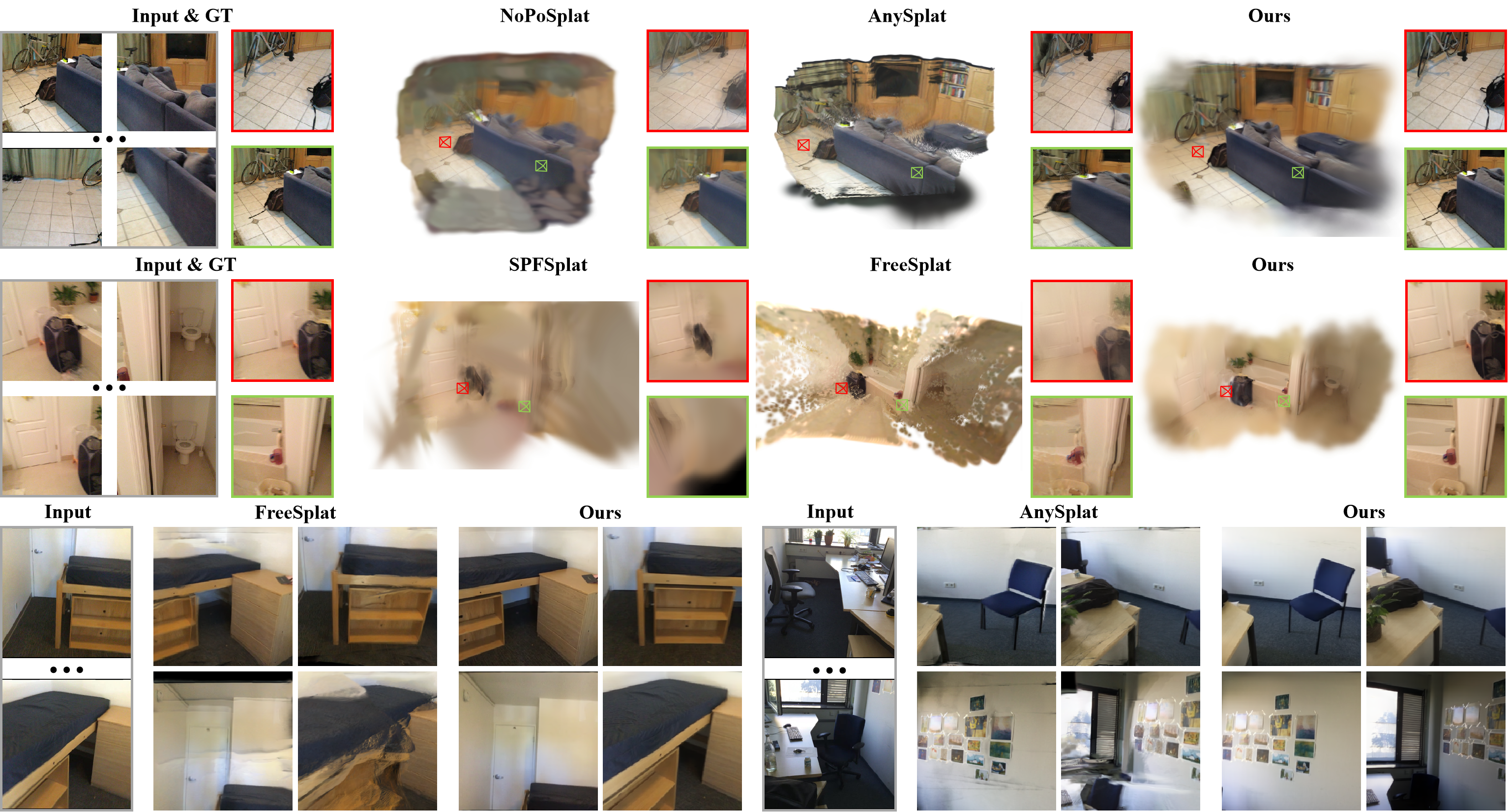}
\caption{Scene-level visualizations and multiple novel viewpoints renderings.}
\label{fig:gaussian_nvs}
\end{figure*}

\noindent\textbf{Cross-Dataset Generalization.}
To assess zero-shot generalization, we train TokenSplat on RE10K and directly evaluate it on ScanNet. As shown in the right halves of both Tab.~\ref{tab:re10k-nvs} and Tab.~\ref{tab:re10k-pose}, which report cross-dataset results, our method surpasses all SOTA approaches on both novel view synthesis and pose estimation. Compared to FreeSplat, TokenSplat achieves up to 0.92/0.51 dB higher PSNR, and reduces ATE by 0.013/0.033 over SPFSplat.
In Fig.~\ref{fig:re10k_cross_scannet_nvs}, TokenSplat preserves coherent geometry and clearly reconstructed scene structures. Notably, fine details such as furniture boundaries are better preserved than in competing methods, demonstrating the effectiveness of our method across unseen scenes. The camera pose head further ensures reliable pose estimation under novel camera configurations, jointly contributing to superior cross-dataset performance.

\noindent\textbf{Scene-level Visualization.}
%In Fig. \ref{fig:gaussian_nvs}, we visualize several multi-view reconstructed Gaussian scenes and present rendering results from multiple novel viewpoints. Anysplat, which directly fuses 3D Gaussians, exhibits obvious unaligned fusion artifacts (top) and fusion cracks between viewpoints (ground area).
% , and the scene is somewhat distorted and difficult to align with the real camera viewpoints
%Freesplat generates a large number of inaccurately scattered Gaussians, while NoPoSplat and SPFSplat have limited scalability and struggle to generalize to viewpoints far beyond those seen during training. In contrast, our method maintains clearer and more accurate reconstructed structures even when extended to viewpoints significantly exceeding the number used in training.
To further illustrate the reconstruction quality, Fig.~\ref{fig:gaussian_nvs} presents scene-level visualizations, where the camera is pulled back to reveal the global 3D Gaussian structures, along with several rendered novel views.
AnySplat, which directly fuses 3D Gaussians, exhibits noticeable misalignment artifacts and visible fusion cracks across viewpoints.
FreeSplat generates numerous scattered Gaussians, while NoPoSplat and SPFSplat show poor scalability and fail to generalize to unseen distant viewpoints.
In contrast, our method preserves coherent and geometrically consistent reconstructions, even when rendered from viewpoints far beyond the training range.

\begin{table}[t!]
\renewcommand\arraystretch{0.85} 
\setlength{\tabcolsep}{1.5pt}
\centering
\caption{Component ablations on RE10K (8 view).}
\resizebox{0.98\columnwidth}{!}{
\begin{tabular}{lcccccc}
\toprule
% \multirow{2}{*}{} & \multicolumn{3}{c}{NVS} & \multicolumn{3}{c}{Pose}\\ 
% \cmidrule(lr){2-7}
Method& PSNR$\uparrow$ & SSIM$\uparrow$ & LPIPS$\downarrow$ & ATE$\downarrow$ & RPE-t$\downarrow$ & RPE-r$\downarrow$ \\
\midrule
(a) \textbf{Ours} & 26.15 & 0.858 & 0.135 & 0.012 & 0.019 & 0.458  \\
\midrule
(b) w Pixel Head & 25.33 & 0.832 & 0.159 & 0.018 & 0.028 & 0.496 \\
(c) w Any. Fusion. & 25.77 & 0.847 & 0.148 & 0.020 & 0.027 & 0.489 \\
(d) w/o ADF-Dec. & 25.88 & 0.845 & 0.146 & 0.023 & 0.029 & 0.504 \\
(e) w/o intrin. emb. & 25.54 & 0.835 & 0.157 & 0.016 & 0.022 & 0.471 \\

\bottomrule
\end{tabular}
}
\label{tab:ablations}
\end{table}

\subsection{Ablation Analysis}
We perform ablation studies on RE10K (8 views), summarized in Tab.~\ref{tab:ablations}. Compared to our full model (a), 
(b) replacing the Token-aligned Gaussian Prediction with a pixel-aligned Gaussian head degrades both reconstruction and pose estimation, with SSIM dropping by 0.026 and RPE-r increasing by 0.038; 
(c) using AnySplat-style Gaussian fusion, which fuses pixel-aligned Gaussians by predicting per-Gaussian fusion confidence, improves over the pixel-aligned head but still underperforms our token-based fusion, with PSNR 0.38 dB lower; 
(d) replacing the ADF-decoder with a standard ViT decoder that concatenates camera and image tokens for attention increases pose entanglement, leading to 0.046 higher RPE-r and 0.011 higher LPIPS (lower is better), highlighting the benefit of directionally constrained communication for disentangling pose and image features;  
(e) removing the known intrinsic camera embedding reduces reconstruction quality, as the model struggles to capture scale, though pose estimation remains competitive.
Overall, these results demonstrate the effectiveness of our model designs for accurate multi-view reconstruction and camera pose estimation.
\section{Conclusion}
We present \textit{TokenSplat}, a feed-forward framework for joint 3D reconstruction and camera pose estimation from unposed multi-view images. By leveraging token-based multi-view fusion and ADF-Decoder, it achieves clean feature disentanglement and stable aggregation, enabling coherent reconstruction without iterative refinement. TokenSplat consistently improves reconstruction and pose accuracy and generalizes well, achieving competitive zero-shot performance when trained on a single RE10K-scale dataset.
\clearpage
{
    \small
    \bibliographystyle{ieeenat_fullname}
    \bibliography{main}
}

\clearpage
\setcounter{page}{1}
\renewcommand\thesection{A\arabic{section}}
\renewcommand*{\thefigure}{A\arabic{figure}}
\renewcommand*{\thetable}{A\arabic{table}}

%% 重新计数
\setcounter{section}{0}
\setcounter{figure}{0}
\setcounter{table}{0}

\maketitlesupplementary
\section{More Implementation Details of Experiment}
\noindent\textbf{More Training Details.}
Our model is implemented using PyTorch, and all models are trained on a single A800 GPU. Comparable results can also be achieved on a single A6000 or A100 GPU by adjusting to a smaller batch size and increasing the number of training iterations. Following strategies similar to those in \cite{ye2024nopo} and \cite{li2025vicasplat}, we train with a batch size of 16, a learning rate of $2\times10^{-5}$ for the backbone, and a learning rate of $2\times10^{-4}$ for the other components for 30,000 iterations under the 2-view settings. Subsequently, for multi-view scenarios, we fine-tune the model for 100,000 iterations with a backbone learning rate of $4\times10^{-6}$, a learning rate of $4\times10^{-5}$ for the other components, and a batch size of 1. For the 28-view test on ScanNet, we use the model trained under the 10-view setting to evaluate the model's ability to generalize to a larger number of views. All other models, except for AnySplat, adopt a similar setup.

\noindent\textbf{More Details of MVSplat and FreeSplat Results.}
All settings follow those specified in the official FreeSplat repository. For the 28-view results of FreeSplat on ScanNet, we use the model trained with the ``fvt" setting (with the maximum number of views during training still set to 10), as this configuration offers better scalability and leads to higher performance for FreeSplat.

\noindent\textbf{More Details of NoPoSplat Results.}
We use the official 2-view checkpoint from the repository and further fine-tune NoPoSplat with the exact same settings as our method to ensure a fair comparison. Additionally, all methods use the same view alignment protocol as NoPoSplat for evaluation to ensure complete fairness.

\noindent\textbf{More Details of SPFSPlat Results.}
According to the official training configuration provided for SPFSplat, 200,000 training iterations are required. We use the official 2-view checkpoint from the repository and further fine-tune them for 200,000 iterations on the multi-view setting to ensure sufficient training of SPFSplat.

\noindent\textbf{More Details of VicsSplat Results.}
For VicaSplat, the official code specifies the training requirements for each step on RE10K, and we strictly follow the training and fine-tuning procedures provided by the official code. For training on ScanNet, we follow our training and fine-tuning settings, and, as specified by the official guidelines, we perform point distillation before training.

\noindent\textbf{More Details of AnySplat Results.}
Since the training dataset of AnySplat is approximately four times larger than the number of scenes in our RE10K training set and 95 times larger than the number of images in ScanNet. Therefore, we report both the zero-shot results of the generic model provided by AnySplat (denoted as AnySplat), as well as the results obtained by training on RE10K and ScanNet following the official instructions (denoted as AnySplat$\ast$). Since AnySplat itself is trained on images with a resolution of 448, using different input resolutions leads to severe artifacts. Therefore, we use 448-resolution inputs for zero-shot testing with AnySplat. It is worth noting that the input image content is kept exactly the same as in other models, with only the resolution being different. For novel view synthesis tests, we render the 3DGS outputs of the models at a resolution of 256×256, ensuring that all models are evaluated at the same resolution. (Rendering at 448 resolution and then resampling to 256 yields even worse results.) Moreover, since AnySplat uses VGGT as its backbone and only supports inputs at 14× resolution, we train AnySplat$\ast$ at a resolution of 266×266, and adopt a testing strategy similar to that of AnySplat.

Results show that while the visual quality of AnySplat is acceptable, its reconstructed scenes exhibit some distortions and deformations. This distortion is not easily noticeable when viewing AnySplat results in isolation, but becomes much more apparent when test views are aligned with ground-truth views for evaluation (following the protocol in NoPoSplat). In many scenes, such distortions are clearly observed after this alignment. Even after fine-tuning on the corresponding scenes, achieving improvements and reaching the performance reported by the official implementation, there remains a significant gap compared to our method.

\section{More Implementation Details of Modulation in Asymmetric Dual-Flow Decoder}

%Inspired by Adaptive Layer Normalization in DiT, we adopt a similar modulation design. Specifically, we use an MLP to predict three modulation parameters (scale, shift, gate) from the camera token at each step. Since our camera tokens are continuously updated within the decoder blocks, after each self-attention, a separate MLP predicts three new modulation parameters (scale, shift, gate) from the updated camera token. 
Inspired by the Adaptive Layer Normalization in DiT, we adopt a similar modulation strategy. Specifically, an MLP predicts three modulation parameters, \textit{i.e.} \textit{scale}, \textit{shift}, and \textit{gate}, from the camera token at each step. Since the camera tokens are continuously updated within the decoder blocks, after self-attention layer, a separate MLP predicts a new set of modulation parameters from the updated camera token.
The normalization parameters (scale and shift) and an additional scaling parameter (gate) are applied as follows: before attention, we perform \( \hat{t}^\mathbf{I}_i \leftarrow  \hat{t}^\mathbf{I}_i \times (1 + \text{scale}) + \text{shift} \); after attention, we update as \( \hat{t}^\mathbf{I}_i  \leftarrow  (1 + \text{gate}) \times \hat{t}^\mathbf{I}_i\).

\section{More Implementation Details of Gaussian Prediction Heads}
\begin{figure}[t!]
\centering\includegraphics[width=0.98\linewidth]{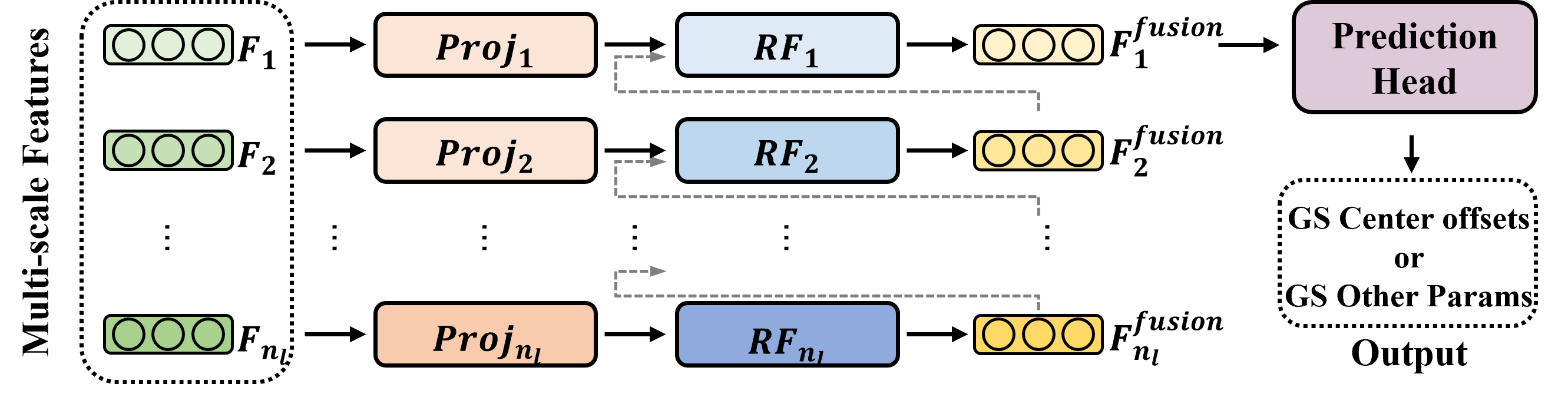}
    \caption{Structure of our Gaussian Prediction Head.}
    \label{fig:tgh}
\end{figure}

In this section, we further elaborate on the detailed design of our Gaussian Prediction heads, as illustrated in Fig. \ref{fig:tgh}. our Gaussian Prediction heads take the multi-scale features from the fused token decoder outputs as input. In our implementation, we follow the DPT head by selecting features from the 0-th, 6-th, 9-th, and 12-th layers of the decoder as inputs $F_1$, $F_2$, $F_3$, $F_4$ to the Gaussian Prediction head.

\begin{figure}[t!]
\centering\includegraphics[width=0.98\linewidth]{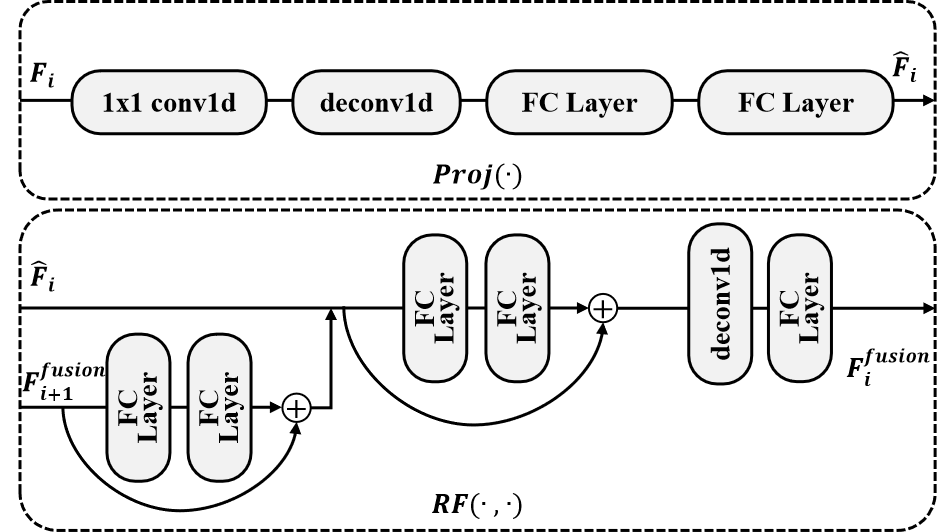}
    \caption{Network details in our Gaussian Prediction Head.}
    \label{fig:tgh-module}
\end{figure}

The multi-scale features are first individually processed by a projection module, denoted as $Proj(\cdot)$, as illustrated in the upper part of Fig. \ref{fig:tgh-module}. This module consists of a 1×1 Conv1D layer for channel transformation, followed by a deconvolution layer for upsampling. Subsequently, two fully connected layers are used for linear mapping to get $\hat{F}_i$. Ultimately, all features are unified to the same number of channels, which is set to 256 in our implementation. 

Subsequently, we perform deep-to-shallow fusion from feature \(\hat{F}_4\) to feature \(\hat{F}_1\), as shown in the lower part of Fig. \ref{fig:tgh-module}. For feature \(\hat{F}_4\), we directly apply a residual module composed of two fully connected layers for further feature transformation. For shallower features \(\hat{F}_i\), we first add the fused feature from the deeper layer \(F_{i+1}^{fusion}\) after passing it through a residual module consisting of two fully connected layers, and then apply another residual module with two fully connected layers for feature transformation. Finally, the output is upsampled using a deconvolution module and further processed by a fully connected layer to obtain \(F_i^{fusion}\).

\section{More Implementation Details of Pose Heads}
The pose head takes the camera token output from the Asymmetric Dual-Flow Decoder as input and consists of activation functions and fully connected layer. Specifically, we use a ReLU activation function and a fully connected layer with dimensions $(768, 7)$ in this work. The pose head outputs camera rotation (as a quaternion) and translation, which are then converted to unit dual quaternions for loss computation. Our pose head is designed with a simple structure, as the Asymmetric Dual-Flow Decoder effectively disentangles camera features, allowing accurate predictions with minimal complexity. 

\begin{figure}[t!]
\centering\includegraphics[width=0.98\linewidth]{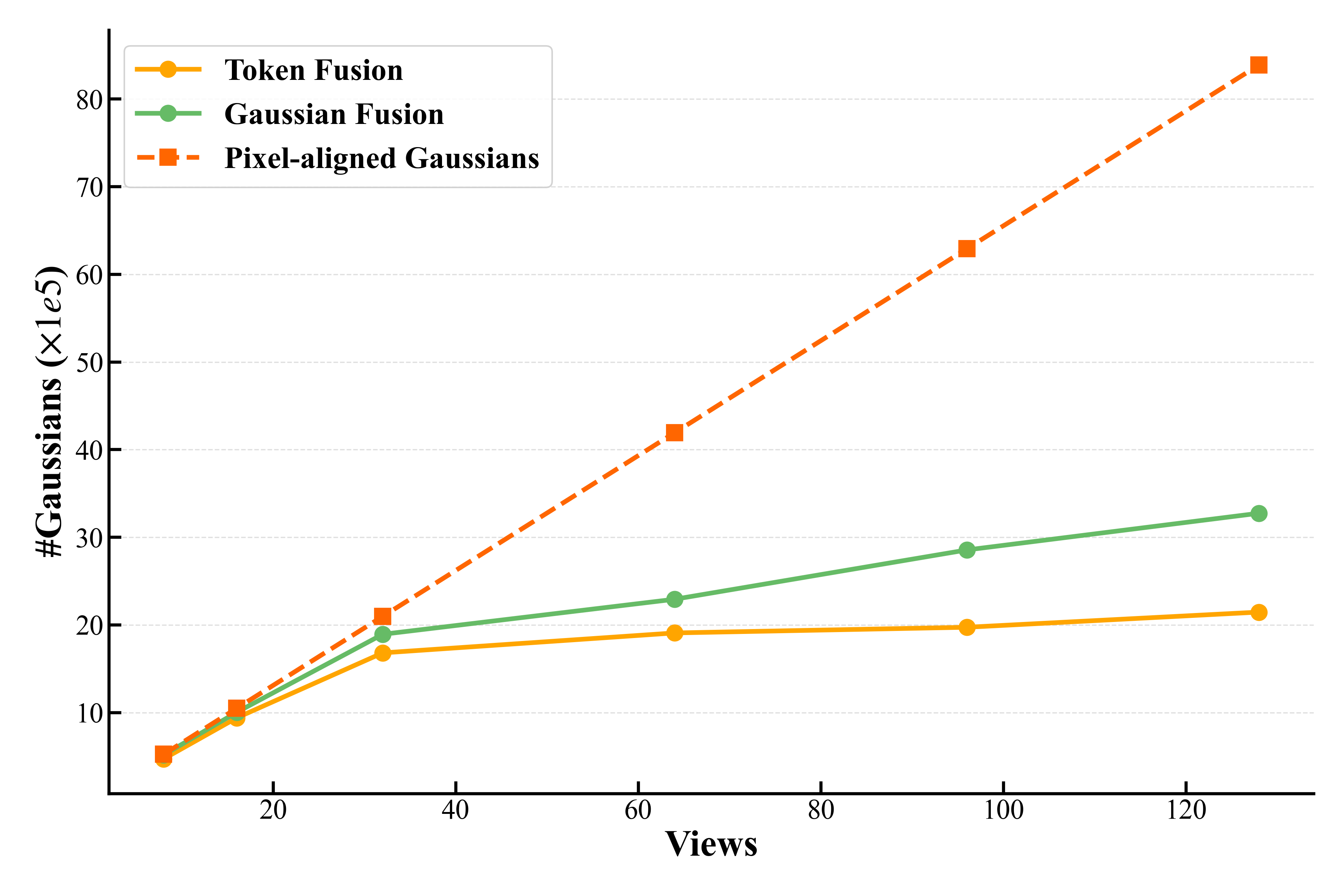}
    \caption{Number of Gaussians with increasing input views for pixel-aligned GS, Gaussian fusion methods, and our Token Fusion operation.}
    \label{fig:supp_gaussian_num}
\end{figure}

\begin{figure}[t!]
\centering\includegraphics[width=0.98\linewidth]{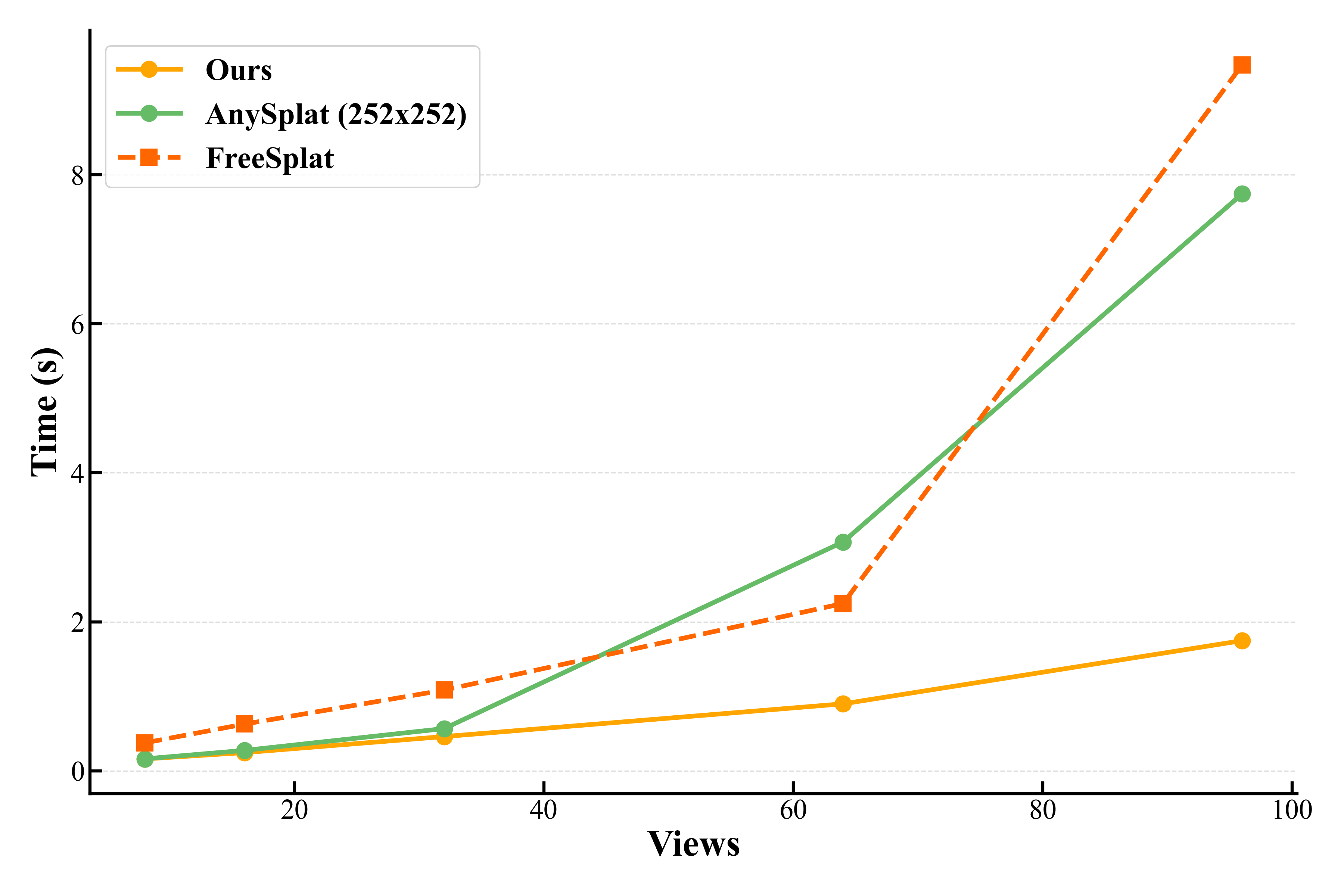}
    \caption{Inference time with increasing number of input views.}
    \label{fig:supp_inference_time}
\end{figure}

\section{More Experimental Analysis}
\subsection{Efficiency}

\begin{figure*}[t!]
\centering\includegraphics[width=0.98\linewidth]{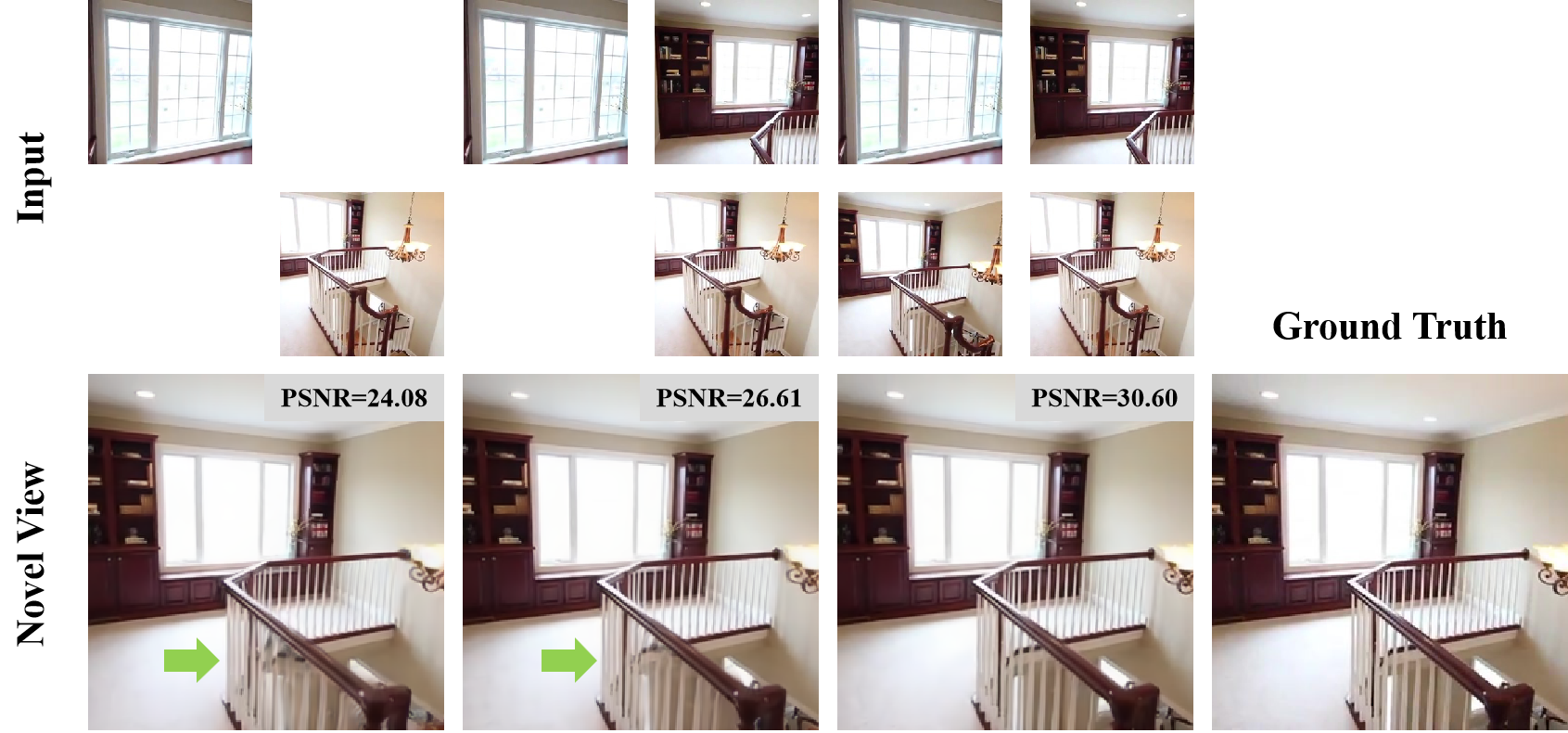}
    \caption{Visualization Results with increasing number of input views.}
    \label{fig:supp_multi_view}
\end{figure*}

We investigate the impact of different fusion methods on the number of Gaussian primitives and show the result in Fig. \ref{fig:supp_gaussian_num}. As the number of input views increases, the number of pixel-aligned Gaussian primitives grows linearly. In contrast, using Gaussian fusion modules in 3D space directly or our Token Fusion operation helps stabilize the growth of Gaussians. It can be observed that as the number of views increases, our Token Fusion operation leads to a much slower growth in the number of Gaussians, making the count more stable compared to Gaussian fusion.
We investigate how the model's inference time changes as the number of input views increases and show the result in Fig. \ref{fig:supp_inference_time}.
As other models experience excessive growth in model size and memory usage with increasing views, we mainly compare with FreeSplat and AnySplat. Since AnySplat only supports up to $14\times$ resolution, we use a resolution of $252\times 252$, while FreeSplat and ours use $256\times 256$. Our method maintains more stable inference time as the number of views increases. This is because, unlike Gaussian fusion methods where computation in the Gaussian prediction stage remains proportional to the number of input images, our approach performs fusion first, fully decoupling the number of Gaussians from the number of images, resulting in more stable inference time.

\subsection{Visualization Results with Increasing Number of Views}
We further demonstrate the impact of increasing the number of input views on novel view synthesis. Starting with the first and last views, we gradually add more intermediate views. As shown in Fig. \ref{fig:supp_multi_view}, additional input views progressively enhance scene completeness and visual details, especially in challenging regions such as railings.

\section{More Visualization Results}
\begin{figure*}[t!]
\centering
\includegraphics[width=0.96\textwidth]{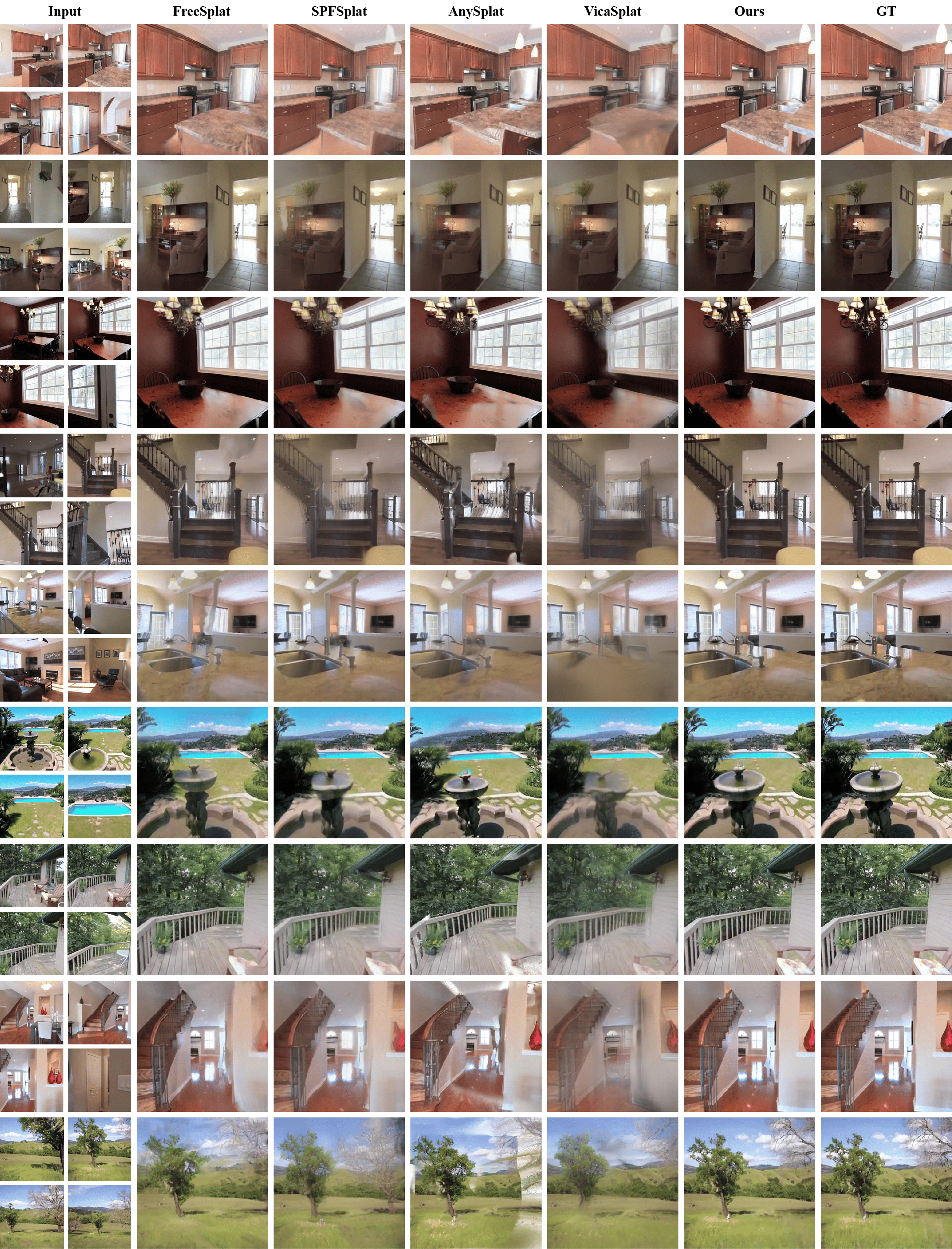}
\caption{More qualitative comparisons on RE10K with 4 input images.}
\label{fig:supp_re10k_4view}
\end{figure*}

\begin{figure*}[t!]
\centering
\includegraphics[width=0.96\textwidth]{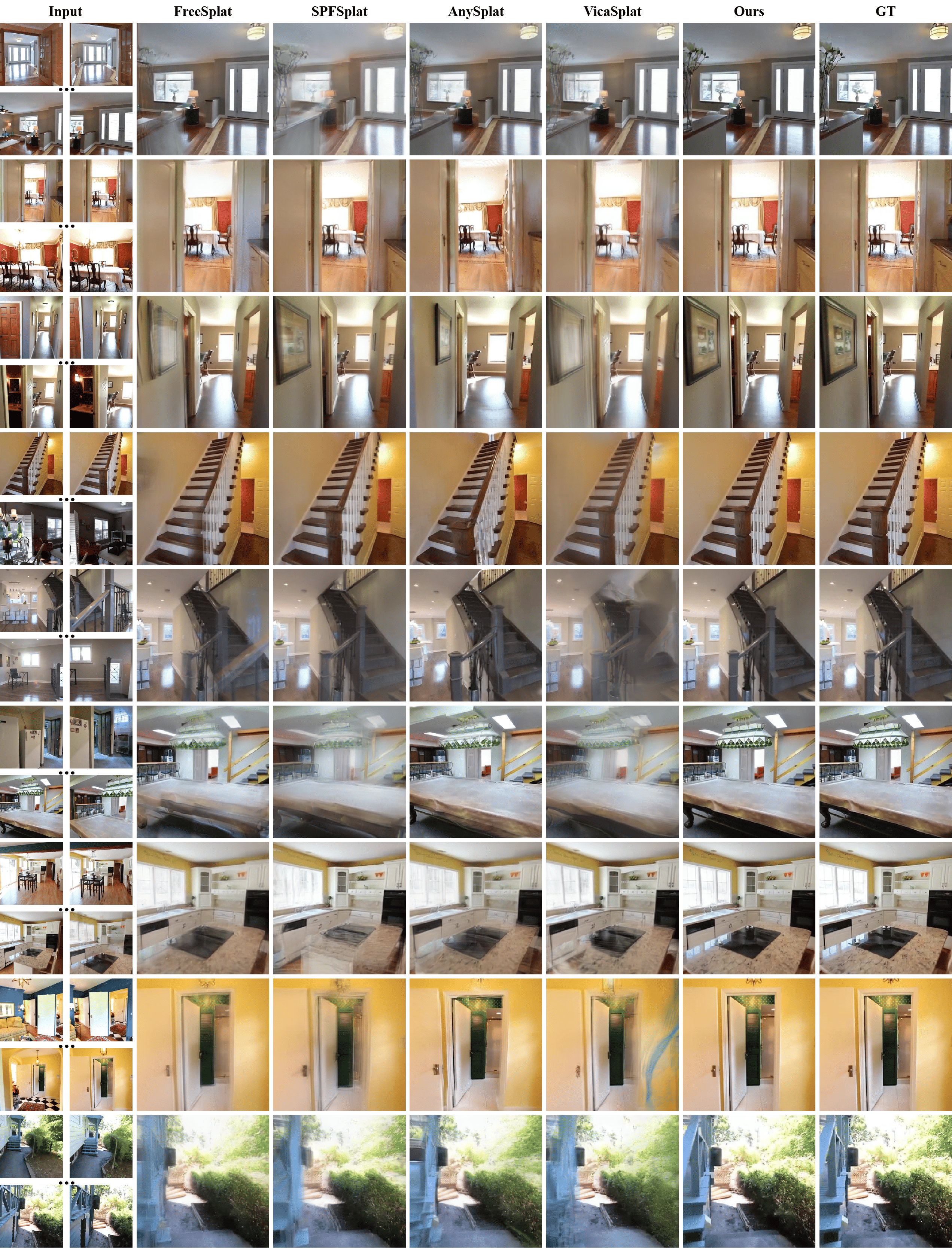}
\caption{More qualitative comparisons on RE10K with 8 input images.}
\label{fig:supp_re10k_8view}
\end{figure*}

\begin{figure*}[t!]
\centering
\includegraphics[width=0.96\textwidth]{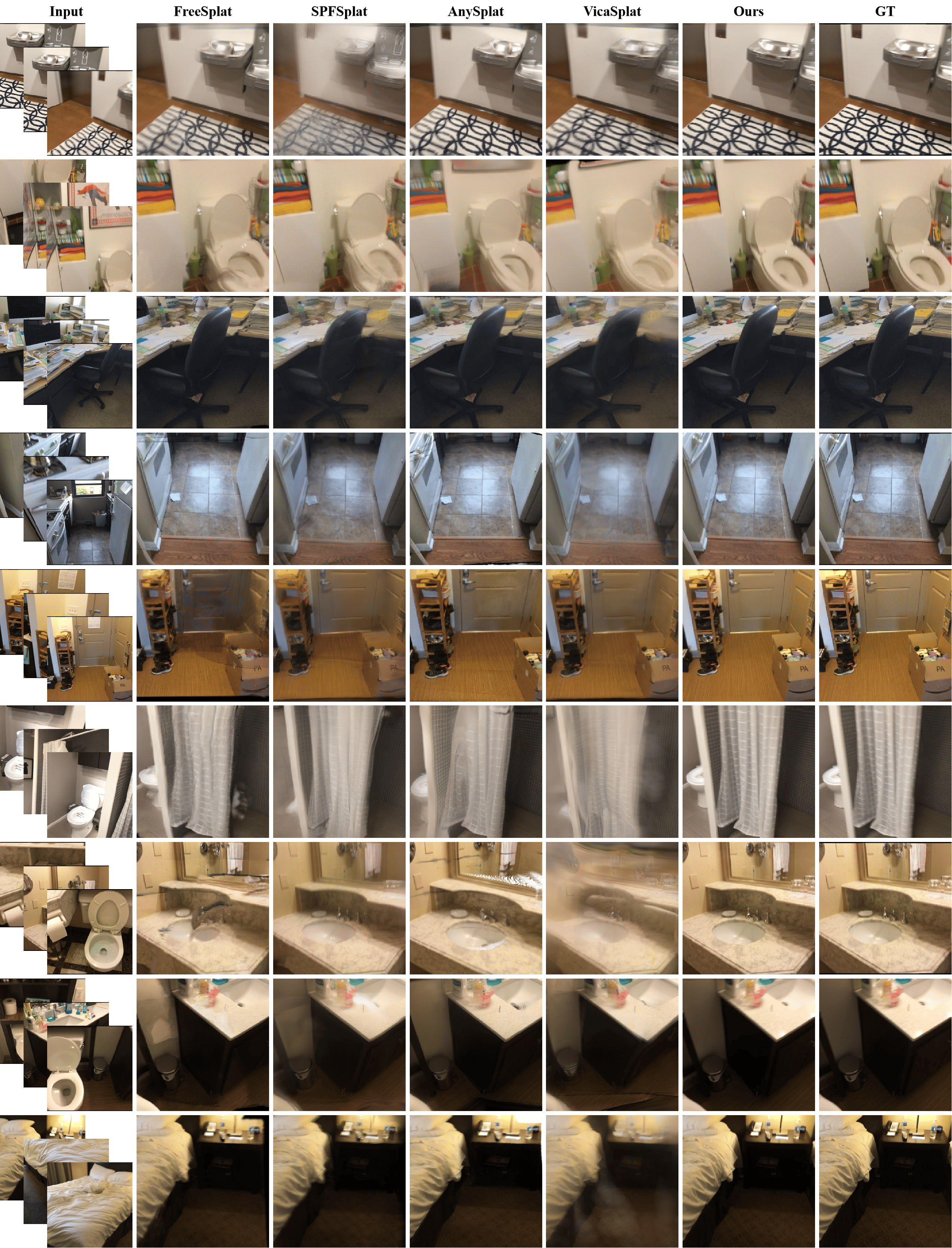}
\caption{More qualitative comparisons on ScanNet with 3 input images.}
\label{fig:supp_scannet_3view}
\end{figure*}

\begin{figure*}[t!]
\centering
\includegraphics[width=0.96\textwidth]{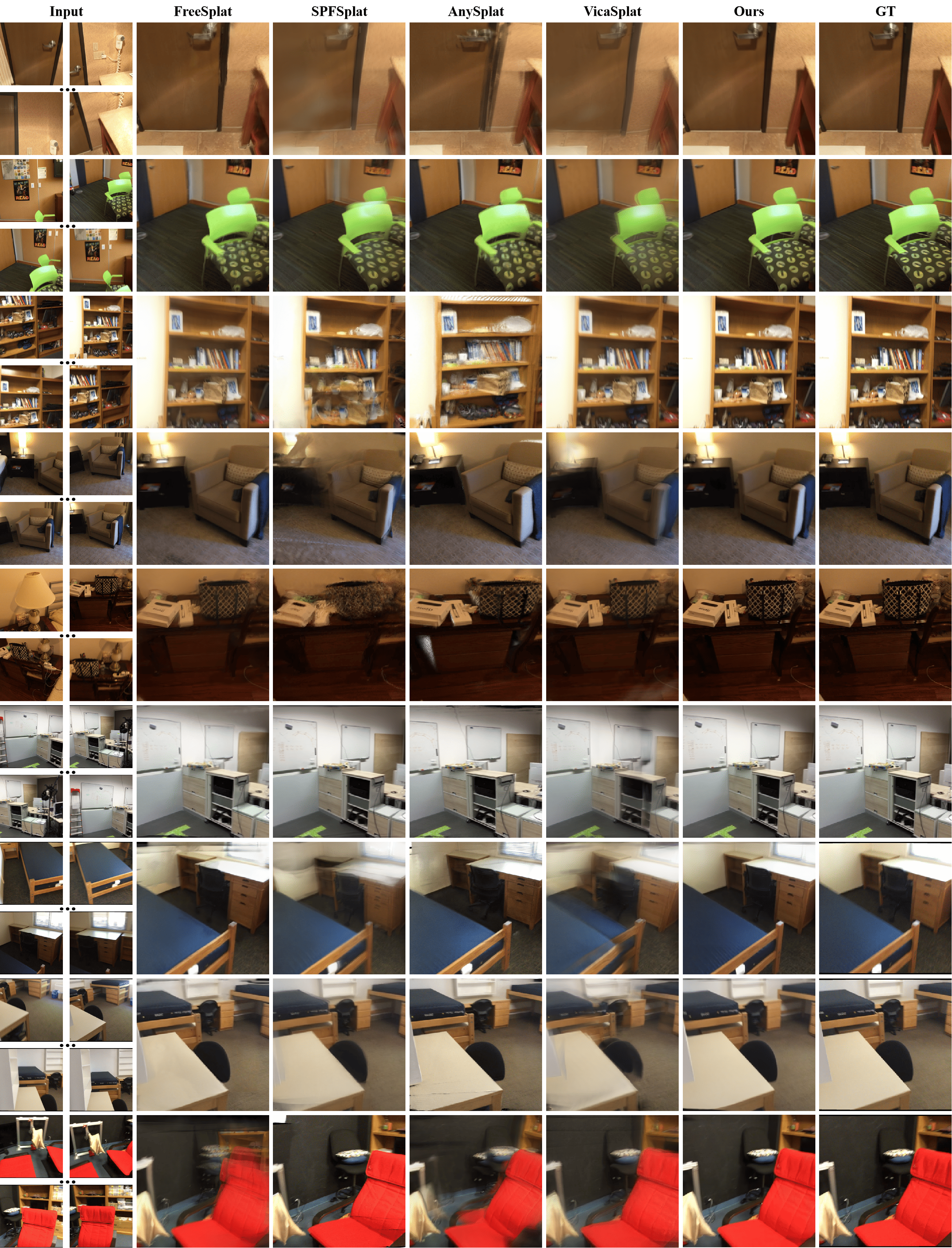}
\caption{More qualitative comparisons on ScanNet with 10 input images.}
\label{fig:supp_scannet_10view}
\end{figure*}

\begin{figure*}[t!]
\centering
\includegraphics[width=0.96\textwidth]{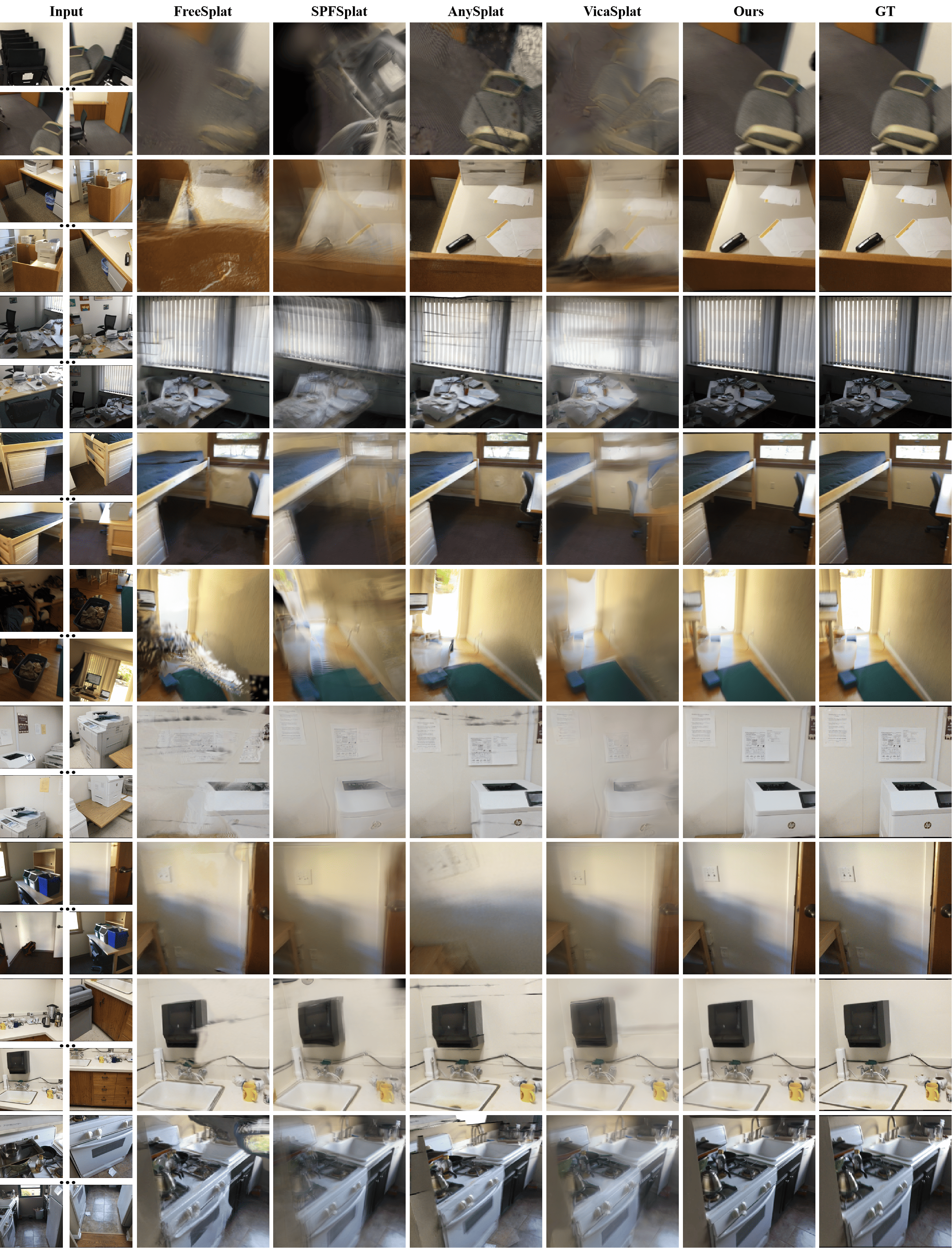}
\caption{More qualitative comparisons on ScanNet with 28 input images.}
\label{fig:supp_scannet_28view}
\end{figure*}

We present additional qualitative comparisons with baselines on the RE10K and ScanNet datasets in Fig. \ref{fig:supp_re10k_4view} to Fig. \ref{fig:supp_scannet_28view}, further demonstrating the effectiveness and improvements of our method. Our approach achieves stable and superior performance across varying numbers of input images and diverse input data.

\section{Video Demo}
The project webpage includes comparison videos between our approach and state-of-the-art methods.
Whether in rendered videos or when observing the entire scene from a distance,
our method consistently exhibits more complete and coherent structures,
whereas existing methods suffer from blurring or defocus-like artifacts,
and some even exhibit severe structural issues.
For example, when running SPFSplat and VicaSplat with more input views than used during training,
we observe significant scene compression at the boundaries, making it difficult to effectively extend outward.
FreeSplat projects Gaussians into 3D space using depth maps, resulting in a large number of floating artifacts.
In the left-side close-up rendering video of AnySplat,
noticeable cracks caused by unsuccessful Gaussian fusion can be observed near the mirror at the beginning,
and multiple issues arising from 3D Gaussian fusion are also clearly visible in the long-range scene.
Furthermore, our method exhibits remarkable generalization ability, as evidenced by the reconstruction results on scenes from different datasets and \textbf{in-the-wild data casually captured with mobile phones}.


\end{document}